\documentclass[article,twocolumn,9pt]{IEEEtran}
\ifCLASSINFOpdf
  \usepackage[pdftex]{graphicx}
  \DeclareGraphicsExtensions{.pdf,.jpeg,.png,.jpg}
\fi
\usepackage[cmex10]{amsmath}
\usepackage{amssymb}
\usepackage{multirow}
\usepackage{booktabs}
\usepackage{cite} % Reorders cites
\ifCLASSOPTIONcompsoc
  \usepackage[caption=false,font=normalsize,labelfont=sf,textfont=sf]{subfig}
\else
  \usepackage[caption=false,font=footnotesize]{subfig}
\fi
\newcommand{\rulesep}{\unskip\ \vrule\ }
\usepackage{hyperref}
\usepackage{etoolbox}
\makeatletter
\patchcmd{\@makecaption}
  {\scshape}
  {}
  {}
  {}
\makeatletter
\patchcmd{\@makecaption}
  {\\}
  {.\ }
  {}
  {}
\makeatother
\DeclareMathSymbol{\shortminus}{\mathbin}{AMSa}{"39}
\newcommand{\minisection}[1]{\smallskip\textbf{#1}}

\begin{document}
\title{Learning to adapt class-specific features across domains\\for semantic segmentation}

\author{
\IEEEauthorblockN{Mikel Menta\IEEEauthorrefmark{1}, Adriana Romero\IEEEauthorrefmark{2}, Joost van de Weijer\IEEEauthorrefmark{1}}\\
\IEEEauthorblockA{\IEEEauthorrefmark{1}Computer Vision Center,
Universitat Aut\`onoma de Barcelona\\
\IEEEauthorrefmark{2}McGill University, Montreal
\thanks{Corresponding author: mikel.menta@cvc.uab.es}
\thanks{Code: \href{https://github.com/mkmenta/domain_adapt_segm}{https://github.com/mkmenta/domain\_adapt\_segm/}}}
}

% The paper headers
\markboth{Master Thesis Dissertation, Master in Computer Vision, %
September 2018}{}%

\maketitle

\begin{abstract}
Recent advances in unsupervised domain adaptation have shown the effectiveness of adversarial training to adapt features across domains, endowing neural networks with the capability of being tested on a target domain without requiring any training annotations in this domain. The great majority of existing domain adaptation models rely on image translation networks, which often contain a huge amount of domain-specific parameters. Additionally, the feature adaptation step often happens globally, at a coarse level, hindering its applicability to tasks such as semantic segmentation, where details are of crucial importance to provide sharp results. In this thesis, we present a novel architecture, which learns to adapt features across domains by taking into account per class information. To that aim, we design a conditional pixel-wise discriminator network, whose output is conditioned on the segmentation masks. Moreover, following recent advances in image translation, we adopt the recently introduced \textit{StarGAN} architecture as image translation backbone, since it is able to perform translations across multiple domains by means of a single generator network. Preliminary results on a segmentation task designed to assess the effectiveness of the proposed approach highlight the potential of the model, improving upon strong baselines and alternative designs.
 
\end{abstract}


\IEEEpeerreviewmaketitle

\section{Introduction}
 \IEEEPARstart{I}{n} the last decade, deep learning has become the \textit{de facto} standard in many machine learning application domains such as computer vision, natural language processing or speech. More specifically, in computer vision, the success of AlexNet \cite{alexnet} in the ImageNet Large Scale Visual Recognition Challenge (ILSVRC) in 2012 changed the research landscape. Since then, deep learning architectures have been quickly spreading and have shown impressive results in tasks such as image classification \cite{alexnet,vgg,inception,resnet,densenet}, semantic segmentation \cite{fcn,pspnet,tiramisu,denseseg,deeplab,dilatednet,unet,skipconn,segnet}, or object detection \cite{frcnn,yolo,yolo3}, among many others.

These recent successes of deep neural networks in many application domains have been attributed, within significant degree, to the availability of large-scale labeled datasets that can be used for training, the increase of computational power through GPUs and the development of sophisticated approaches. However, even if these networks are generally able to achieve a high performance in unseen data samples from the same dataset, they often don't generalize well to new datasets and tasks. This problem is referred to as domain shift or bias. A typical procedure to tackle the domain shift across different datasets is to fine-tune the network trained on a dataset, with samples from the new dataset or task. However, this procedure usually requires having access to a considerable amount of additional annotated data. 

Furthermore, tasks such as semantic segmentation especially suffer from an extremely expensive annotation, due to the pixel-wise nature of the task. To mitigate labeling efforts, many works suggest exploiting datasets built from data generated by computer simulated environment, such as GTA5 \cite{gta5} and SYNTHIA \cite{synthia}, where infinite amounts of data can be easily collected. Unfortunately, training on simulated data leads to the above-mentioned domain shift, when applying the trained models to a real environment, achieving poor generalization. 

Motivated to overcome the domain shift problem, domain adaptation methods propose a handful of approaches to improve the knowledge transfer from a source domain to a target domain. Nevertheless, in the extreme setting of unsupervised domain adaptation, the knowledge needs to be adapted without any target domain annotation. Recent advances in unsupervised domain adaptation mostly target the task of image classification \cite{unsudomadapt, drcn, labelefficient, dirtt, adda}, with several extensions to other tasks such as semantic segmentation \cite{can-structuredda, domadaptgener, structuredadapt} and object detection \cite{rcnn, weakdetect}.

The majority of the recent domain adaptation literature tries to obtain invariant representations between both source and target domains through adversarial training, following \cite{unsudomadapt}. In this adversarial game, generally, the discriminator network aims to distinguish the original domain of the feature representation, while the feature encoding network intends to extract indistinguishable representations from both domains by fooling the discriminator. It is worth mentioning that this adversarial cross-domain feature matching is broadly performed at a coarse level, in a global way \cite{can-structuredda, domadaptgener, structuredadapt, rcnn, weakdetect}, without paying special attention to fine-grained information such as object positions/sizes or, more generally, class distributions, which are relevant for tasks such as semantic segmentation. Note that extracting such detailed information increases the challenges of domain adaption per se.

Additionally, many approaches incorporate image translation or reconstruction objective to obtain more general features, which may be composed of separate networks to process each domain. These auxiliary objectives often introduce additional modules, leading to a vast amount of parameters, which are then discarded during the testing phase.

In this thesis, we present a novel unsupervised domain adaptation model for semantic segmentation, which addresses the previous concerns. The contributions are twofold:
\begin{enumerate}
    \item We successfully reduce the amount of parameters of our domain adaptation network to roughly the half w.r.t. many other methods, by following the idea of \textit{StarGAN} of providing an additional channel to the input of the network with the corresponding domain label. 
    \item We present a novel pixel-wise discriminator training procedure that is able to perform a local discrimination of the feature representations by taking into account per class information. This is achieved by applying the segmentation predictions to the discriminator output, allowing the network to perform the feature distribution matching at class level, instead of globally.
\end{enumerate}

To assess the impact of the contributions of this thesis, we modify MNIST dataset to simulate characteristics that we encounter in real scenarios, namely differences in  appearance (texture, color, brightness) between different domains, as well as in per class pixel distributions. We report both qualitative and quantitative results in those MNIST modifications as a proof of concept of the method proposed in this thesis and show that our local class specific feature matching significantly outperforms the commonly employed global feature matching, as well as the baselines. We also provide an ablation study of our model from the number of parameters in the model's backbone perspective, in addition to the adversarial feature matching, outlining the benefits of our contribution.

The rest of the thesis is organized as follows. Section~\ref{sec:sota} reviews the recent state-of-the-art literature. Then, Section~\ref{sec:back} introduces the concepts needed for the understanding of this thesis. After that, Section~\ref{sec:method} details the proposed approach. Section~\ref{sec:exp} motivates the experiments and comparisons w.r.t. baselines and ablated models, Section~\ref{sec:res} reports the obtained qualitative and quantitative results, and finally, Section~\ref{sec:concl} draws the conclusions of this thesis and presents potential future research directions.

\section{State of the art}
\label{sec:sota}

This section aims to give the reader an overview of recent state-of-the-art approaches that tackle problems relevant to this thesis, namely image translation, semantic segmentation and unsupervised domain adaptation, with special focus on the latter.

\subsection{Image translation}
\label{sec:image-translation}
Image translation and style transfer consist on transferring the appearance of a certain reference image or domain of images to a target image. These areas have been benefiting from a large amount of contributions in recent years. In works like~\cite{perceptual,style,instancenorm,textnets}, approaches are based on information extracted from increasingly abstract layers of a pretrained network. The goal can be summarized as trying to obtain a low level style representation similar to the reference image, while preserving more abstract content-based features.

Nevertheless, with the introduction of Generative Adversarial Networks (GANs)~\cite{gans} and conditional GANs~\cite{cgans}, new research directions exploiting and extending these architectures to perform image translation have emerged \cite{cyclegan, unsupgan, transcan}. Although adversarial architectures have been enjoying increasing popularity among researchers, alternatives based on the combination of features extracted from encoder-decoder architectures have also shown competitive results \cite{imcrossdom}.

A possible way of categorizing image translation approaches is by dividing them into: paired case, where for each input image a corresponding one in the opposite domain is provided; and unpaired case, where there is not such correspondence between images of both domains.

In the paired case, we find works such as \cite{transcan, imcrossdom}, among others. On one hand, \cite{transcan} proposed to generate the translated image with a GAN conditioned on an input image, guiding the network with its corresponding paired image. On the other hand, \cite{imcrossdom} tries to obtain a disentangled representation of the image with both a domain agnostic and a domain specific part. To do so, they train an encoder and a decoder per domain using an aggregation of several loss components.

In the unpaired case, approaches are slightly different. For example, the solution provided in \cite{cyclegan} (explained with more details in Section~\ref{sec:back-cycgan}) is to learn translation mappings among domains (from domain A to domain B and viceversa) by means of adversarial training, and by introducing a cycle-consistency to enforce that, e.g. the translation of an image from A to B, followed by the translation from B to A, leads back to the original image.  Similarly, in \cite{stargan} they propose to use a single generator network, together with a discriminator, to perform translations across multiple domains at the expense of adding a few conditional parameters (detailed in Section~\ref{sec:back-stargan}). Finally, in \cite{fader} the strategy is focused on obtaining representative features that are agnostic to the domains and that, as a result, can be decoded to any target domain

\subsection{Semantic segmentation}
\label{sec:sota-segm}
The task of semantic segmentation, which consists on classifying each pixel of an image according to some known labels, is currently mainly addressed by Deep Neural Networks (DNN) in the state-of-the-art. Current DNN models to tackle pixel-prediction problems are based on Fully Convolutional Networks (FCNs) \cite{fcn}. FCNs endow Convolutional Neural Networks (CNNs) with an upsampling path to recover the input resolution. These networks can be trained end-to-end.

Some of the contributions in this area have focused on obtaining structurally better representations by keeping the pooling indices of the downsampling steps to perform a better upsampling \cite{segnet}, adding skip connections to preserve low level spatial features \cite{unet,skipconn} or combining features at different spatial resolutions \cite{pspnet,tiramisu,denseseg}. Other works have tried to overcome the loss of spatial resolution introduced by pooling layers by means of dilated convolutions to enlarge the receptive field \cite{deeplab,dilatednet} while others have analyzed the benefits and limitations of different receptive field enlargement operations \cite{denseseg}.

Unluckily, an important issue of the semantic segmentation task is the high cost of obtaining labeled data and the big amounts of data required to train high capacity deep learning models. A proposed solution for this problem has been to use weak annotations, which can be obtained in an easier (and less expensive) manner \cite{pointseg,scribbleseg,bboxseg}. Other proposed alternatives involve training the models on synthetic datasets such as SYNTHIA \cite{synthia} or GTA5 \cite{gta5}, where labels can be extracted automatically. Unfortunately, models trained on synthetic datasets tend to not generalize well to real world scenarios.

\subsection{Unsupervised domain adaptation} 
\label{sec:sota-domadapt}
Domain adaptation methods aim to address the challenges posed by existing dataset shifts (e.g. when training and testing samples come from different distributions) with the goal of transferring the knowledge learned in a source domain to a target domain. A well known approach for this problem is training in the source domain and \textit{fine-tuning} in the target domain. However, in many real cases, the ground truth information for the target domain might not be available (\textit{unsupervised domain adaptation}) or might be too sparse (\textit{semi-supervised domain adaptation}). 

Due to the usefulness of solving the unsupervised case, we can find a handful of computer vision work in this area. The majority of them focus or experiment with the \textit{classification} task \cite{dirtt, unsudomadapt,  labelefficient, residualtrans,  drcn,  domsep,  adda,  cada,  walker,  similarity,  labelefficient, partialda, attention}. But we can also find a significant variety of approaches in semantic segmentation \cite{can-structuredda, domadaptgener, structuredadapt}, object detection \cite{rcnn, weakdetect} and depth estimation \cite{monocular-depth, unsupgan}, among others.

The methodologies developed to tackle domain adaptation can vary on some specific aspects depending on the particularities of the task. However, a significant number of works rely on state-of-the-art image translation methods (reviewed in subsection~\ref{sec:image-translation}) or simple image reconstruction with different purposes. Some of these purposes are: (1) obtaining unsupervised features, reconstructing the input images, to perform the task directly \cite{drcn}; (2) translating source images to the target domain and training a network from those images in a supervised manner  \cite{monocular-depth,unsupgan} or (3) translating the source images and using them in a subsequent fine-tuning step \cite{weakdetect}.
Generally, in many cases, the image translation and/or reconstruction tasks are used as additional signal to learn more representative features for the target domain \cite{domsep,domadaptgener}.

It is worth mentioning that the vast majority of approaches to (unsupervised) domain adaptation focus on \textit{matching distributions} of source and target domains at feature level. To do so, state-of-the-art approaches rely on (1) maximizing the correlation between features of both domains \cite{attention}, (2) associating the target samples with the source samples \cite{walker} or (3) obtaining a clustered distribution in the features of the target domain \cite{dirtt}. Nonetheless, a vast set of the recent literature focuses on matching distributions of source and target domains via \textit{adversarial matching} of the features, either by means of a domain discriminating classifier, a \textit{Gradient Reversal Layer} (first introduced in \cite{unsudomadapt}) or the vanilla adversarial objective introduced in \cite{gans}. Some of these approaches propose to learn translated features from source to target domains following the image translation paradigm \cite{can-structuredda}. Moreover, the feature matching can also be done at different depth levels of the network simultaneously \cite{labelefficient, structuredadapt}. A complementary view is to isolate the domain agnostic features from the domain specific ones as in \cite{domsep} using different network modules. In \cite{adda} the authors analyze the previously mentioned alternatives along with an overview of general challenges encountered in domain adaptation. Similarly, in \cite{domadaptgener} the authors exploit and combine several of the previously mentioned components into a single model to obtain better training signals. Recently, works on modeling the domain adaptation problem as a similarity learning task have also emerged in the literature \cite{similarity}. Finally, alternative research directions include (1) predicting the weights of the target model \cite{residualtrans}; (2) addressing the well known mode collapse problem in adversarial training with conditioning \cite{cada}; (3) dealing with a partial unsupervised domain adaptation formulation \cite{partialda}; or (4) addressing and analyzing less popular tasks in the domain adaptation literature, such as object detection \cite{rcnn}.

\section{Background}
\label{sec:back}
In this section, we will go through the design and characteristics of some models needed for the understanding of this thesis. In the following subsections, we will explain fully convolutional network architectures for semantic segmentation \cite{unet,fcn,skipconn,tiramisu} as well as the \textit{CycleGAN} \cite{cyclegan} and \textit{StarGAN} \cite{stargan} architectures for unpaired image translation.

\subsection{Fully convolutional network for semantic segmentation}
\label{sec:back-fcnss}
As mentioned in Section~\ref{sec:sota-segm}, a common approach for tackling the semantic segmentation problem is the use of FCNs that allow to obtain a segmentation mask directly from an image in an end-to-end way. FCNs are composed of a downsampling path (with convolutional and subsampling operations) followed by an upsampling path (with transposed convolutions), which recovers the input resolution. In this thesis, we also use an FCN for performing the segmentation task.

\begin{figure}[ht]
    \centering
    \includegraphics[width=0.7\linewidth]{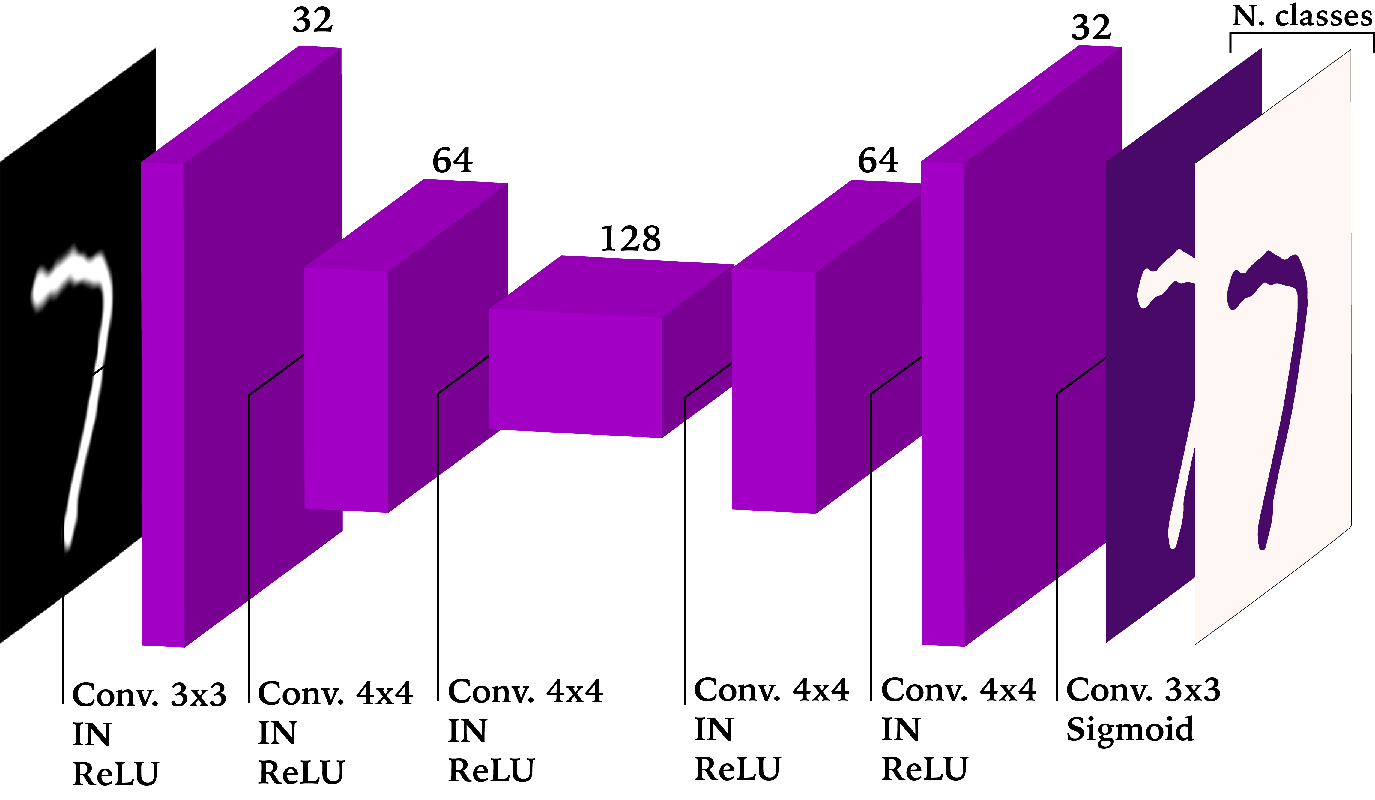}
    \caption{Architecture of the fully convolutional network for semantic segmentation defined in Section~\ref{sec:back-fcnss}. \textit{IN} indicates instance normalization.}
    \label{fig:segm}
\end{figure}

Given a dataset $\mathcal{D}=\{(x^{(i)}, y^{(i)})\}_{i=1}^N$ where $(x^{(i)},y^{(i)})$ are pairs of image and segmentation masks respectively, we define a FCN $S$ which approximates as well as possible the mapping of an input sample $x$, potentially not included in $\mathcal{D}$, to its corresponding label $y$. Figure~\ref{fig:segm} depicts an FCN architecture with an initial convolutional block, 2 downsampling blocks, followed by 2 upsampling blocks and a final convolution that performs the segmentation. Extra details of the implementation and training are explained in Appendix~\ref{apx:implementation}. As will be later explained in the thesis, this model is used as a baseline (Section~\ref{sec:exp:base}) and its decoder, composed of the upsampling path exclusively, is used as segmenter module in our model (Section~\ref{sec:method}).

In the binary segmentation case, we can train $S$ by minimizing the \textit{soft IOU loss} \cite{softiou,skipconn} between the predictions $\hat{y}=S(x)$ and ground-truth segmentations $y$. Considering binary segmentations, where $y^p$ is $1$ if pixel $p$ in segmentation $y$ corresponds to foreground class and $0$ otherwise, the segmentation loss $\mathcal{L}_{segm}$ is defined as follows:
\begin{equation}
    \mathcal{L}_{segm} = \mathbb{E}_{x}\!\!\left[
    1-\frac
        {\sum_{p} \hat{y}^p \cdot y^p}
        {\sum_{p} \left( \hat{y}^p + y^p
        -\hat{y}^p \cdot y^p \right)} 
    \right]. \label{eq:segm:siou}
\end{equation}
Note that for multi-class segmentation problems, the cross-entropy loss is used as de facto standard. Considering the segmentations in a \textit{one-hot} encoding, where $y^{c,p}$ is $1$ if pixel $p$ in segmentation $y$ corresponds to class $c$ and $0$ otherwise, the cross-entropy loss is defined as follows:
\begin{equation}
    \mathcal{L}_{segm} = \mathbb{E}_{x,p,c} \left[ 
        \shortminus y^{c,p} \cdot \log \left(\hat{y}^{c,p}\right) 
        - \left(1 \shortminus y^{c,p}\right) \cdot \log \left(1 \shortminus \hat{y}^{c,p}\right)
    \right].
    \label{eq:segm:ce}
\end{equation}

\subsection{CycleGAN}
\label{sec:back-cycgan}
One of the most popular image translation architectures based in conditional adversarial models is \textit{CycleGAN} \cite{cyclegan}. \textit{CycleGAN} is designed to perform unpaired image translations between two domains of images by training two FCN generators and two discriminators. Each generator is trained to translate an image from one domain to the other.

\begin{figure*}[ht]
    \centering
    \begin{tabular}{c}
         \includegraphics[height=3.5cm]{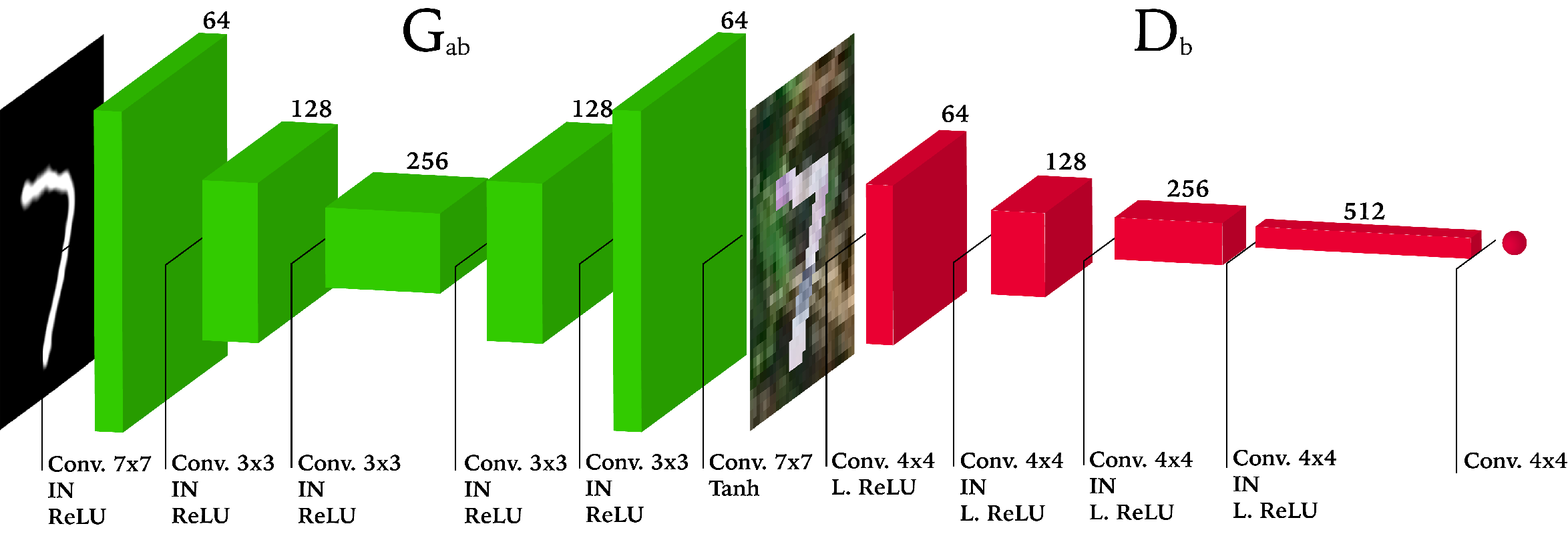} \\
         \includegraphics[height=3.5cm]{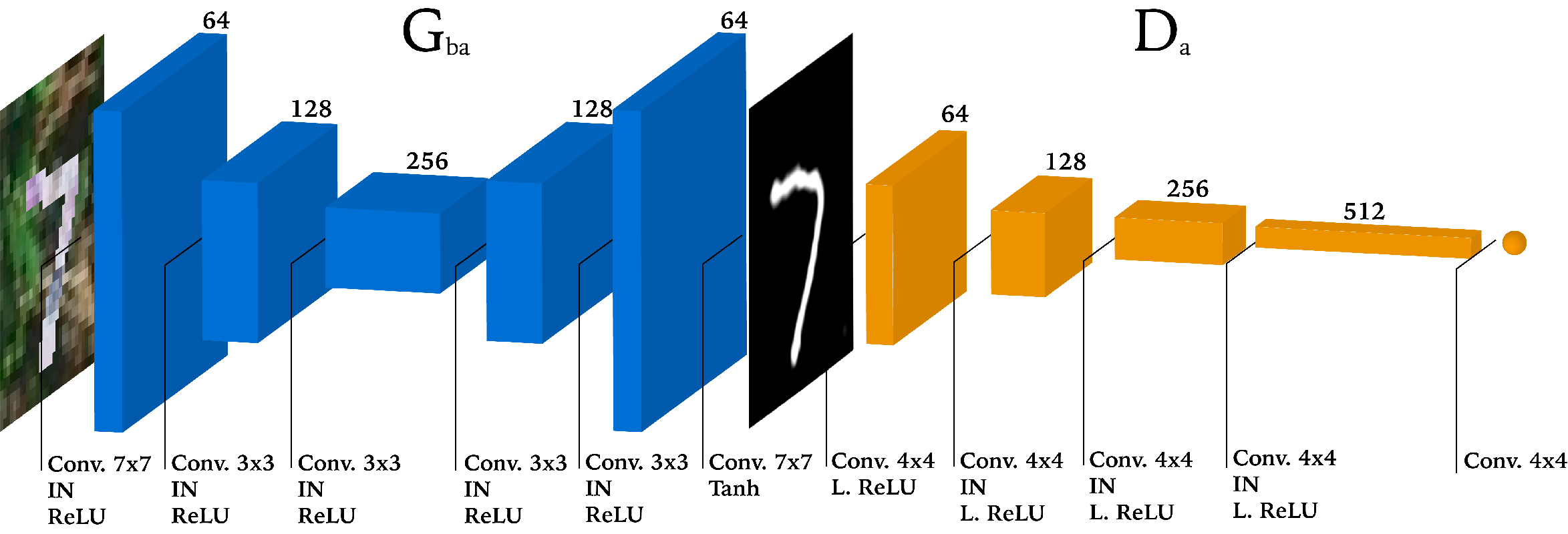}
    \end{tabular}
    \caption{Architecture of the \textit{CycleGAN} \cite{cyclegan} presented in Section~\ref{sec:back-cycgan}. \textit{IN} indicates instance normalization and \textit{L. ReLU} leaky rectified linear unit.}
    \label{fig:cyclegan:adv}
\end{figure*}

Formally, given two datasets of images $\mathcal{D}_a=\{x_a^{(i)}\}_{i=1}^{N_a}$ and $\mathcal{D}_b=\{x_b^{(i)}\}_{i=1}^{N_b}$ sampled from two different domains $a$ and $b$, we define two generator networks $G_{ab}: x_a \rightarrow x_b$ and $G_{ba}: x_b \rightarrow x_a$. Apart from those, we define a discriminator per domain $D_a$ and $D_b$ which aim to distinguish images sampled from their corresponding dataset from the ones generated by $G_{ab}$ and $G_{ba}$. The generator networks are FCNs which, in our implementation, have an initial convolutional block, followed by 2 downsampling blocks, 2 upsampling blocks and a final convolution that maps back to the image domain. The discriminators $D$ follow the \textit{PatchGAN} architecture \cite{transcan} with 4 downsampling blocks and a fully connected layer with a single output. These networks are depicted in Figure~\ref{fig:cyclegan:adv} and overall, they are trained following two objectives:

\minisection{Adversarial losses.} In order to obtain realistic translations for each domain, an adversarial loss is optimized between each pair of generator and discriminator networks. Following the traditional adversarial setting, the discriminator tries to distinguish if an image comes from a real dataset or has been generated, whereas the generator tries to fool the discriminator. Under this idea and, using the Least Squares GAN loss \cite{lsgan}, the following loss components are defined for images from domain $a$:
\begin{eqnarray}
    \mathcal{L}_{adv}^{D_a}\!\!\!\!&=&\!\!\!\!
        \mathbb{E}_{x_a}\!\!\left[ 
            ( D_a(x_a) \shortminus 1 )^2
        \right] 
        +\mathbb{E}_{x_b}\!\!\left[ 
            D_a( G_{ba}(x_b) )^2
        \right] \\[1ex]
    \mathcal{L}_{adv}^{G_{ba}}\!\!\!\!&=&\!\!\!\!
        \mathbb{E}_{x_b}\!\!\left[ 
            (D_a(G_{ba}(x_b)) \shortminus 1)^2
        \right]
\end{eqnarray}
and analogously, the following are defined for images from domain~$b$:
\begin{eqnarray}
    \mathcal{L}_{adv}^{D_b}\!\!\!\!&=&\!\!\!\!
        \mathbb{E}_{x_b}\!\!\left[ 
            ( D_b(x_b) \shortminus 1 )^2
        \right] 
        +\mathbb{E}_{x_a}\!\!\left[ 
            D_b( G_{ab}(x_a) )^2
        \right] \\[1ex]
    \mathcal{L}_{adv}^{G_{ab}}\!\!\!\!&=&\!\!\!\!
        \mathbb{E}_{x_a}\!\!\left[ 
            (D_b(G_{ab}(x_a)) \shortminus 1)^2
        \right]
\end{eqnarray}
The setting of the subnetworks to compute the above-mentioned loss components are represented in Figure~\ref{fig:cyclegan:adv}.

\begin{figure}[ht]
    \centering
    \includegraphics[width=0.7\linewidth]{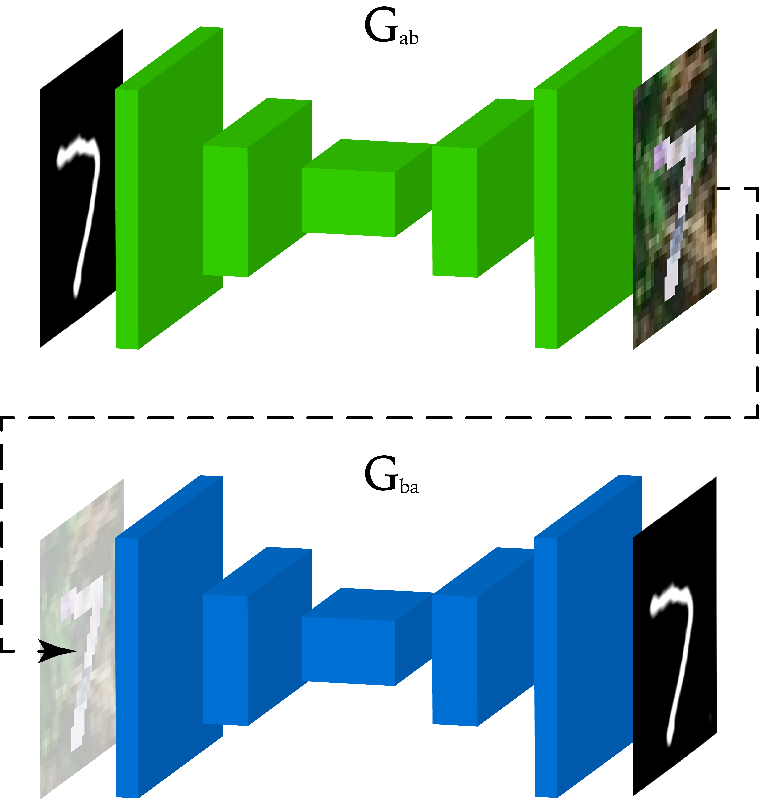}
    \caption{Cycle configuration of the modules in \textit{CycleGAN} \cite{cyclegan} in order to compute $\mathcal{L}_{cyc}$ for domain $a$.}
    \label{fig:cyclegan:cyc}
\end{figure}

\minisection{Cycle consistency loss.} The adversarial loss components don't encourage the model to keep the content intact in the generated translations, they only focus on producing images that look realistic in the translated domain. In order to motivate this behaviour, a cycle consistency loss is added, so that an image translated to the opposite domain and back to the original domain remains invariant. The cycle consistency loss is defined as:
\begin{equation}
\begin{split}
    \mathcal{L}_{cyc} = &\;
        \mathbb{E}_{x_a}\!\!\left[ \left\lVert G_{ba}(G_{ab}(x_a)) - x_a \right\rVert_{1} \right] \\
        &+\mathbb{E}_{x_b}\!\!\left[ \left\lVert G_{ab}(G_{ba}(x_b)) - x_b \right\rVert_{1} \right].
\end{split}
\end{equation}
An example of the network setting in one of the two possible cycles is represented in Figure~\ref{fig:cyclegan:cyc}. Note that an analogous pipeline is used to compute the cycle loss for the remaining domain.

Finally, the model is trained by making the generators $G_{ab}$ and $G_{ba}$ minimize the objective
\begin{equation}
    \mathcal{L}^G = 
    \mathcal{L}_{adv}^{G_{ab}} + \mathcal{L}_{adv}^{G_{ba}} + 
    \lambda \cdot \mathcal{L}_{cyc} \ ,
\end{equation}
where $\lambda$ is an hyper-parameter controlling the relative importance of the two objectives, and making the discriminators $D_a$ and $D_b$ minimize the objective
\begin{equation}
    \mathcal{L}^D = 
    \mathcal{L}_{adv}^{D_{a}} + \mathcal{L}_{adv}^{D_{b}} \ .
\end{equation}

\subsection{StarGAN}
\label{sec:back-stargan}
\textit{StarGAN} \cite{stargan} is a recently proposed alternative to \textit{CycleGAN} to address image translation, and is also the model that we choose as backbone for the unsupervised domain adaptation pipeline introduced in Section~\ref{sec:method}. \textit{StarGAN} offers several practical advantages w.r.t. \textit{CycleGAN}, namely it is designed to perform translations across multiple domains and uses a single generator network and a single discriminator network. To do so, the model expects to receive the translation domain label together with the image to be translated as input to the generator network, performing a different translation according to this label.

Given that in the problem of domain adaptation only a source and a target domain are taken into account, we simplify the notation of the equations to the case of having only two domains. So, considering the datasets $\mathcal{D}_a$ and $\mathcal{D}_b$ introduced in Section~\ref{sec:back-cycgan}, we define a generator network $G$ that, given an image $x \in \{ \mathcal{D}_a \cup \mathcal{D}_b \}$ with its corresponding domain label $l \in \{a,b\}$ and opposite label $\bar{l} \in \{a,b\}$ with $\bar{l} \neq l$, is able to perform a translation $\tilde{x}_{\bar{l}} = G(x_l, \bar{l})$. The translation label $\bar{l}$ is replicated along the spatial dimension to match the size of $x$. We also define a discriminator $D$ that, given an image $x$, predicts (a) whether $x$ is sampled from one of the datasets or has been generated by $G$ (we will denote this prediction as $D_{rf}(x)$) and (b) the domain label of $x$ (we will denote this prediction as $D_{dom}(x)$).

\begin{figure*}[ht]
    \centering
    \begin{tabular}{c}
         \includegraphics[height=3.7cm]{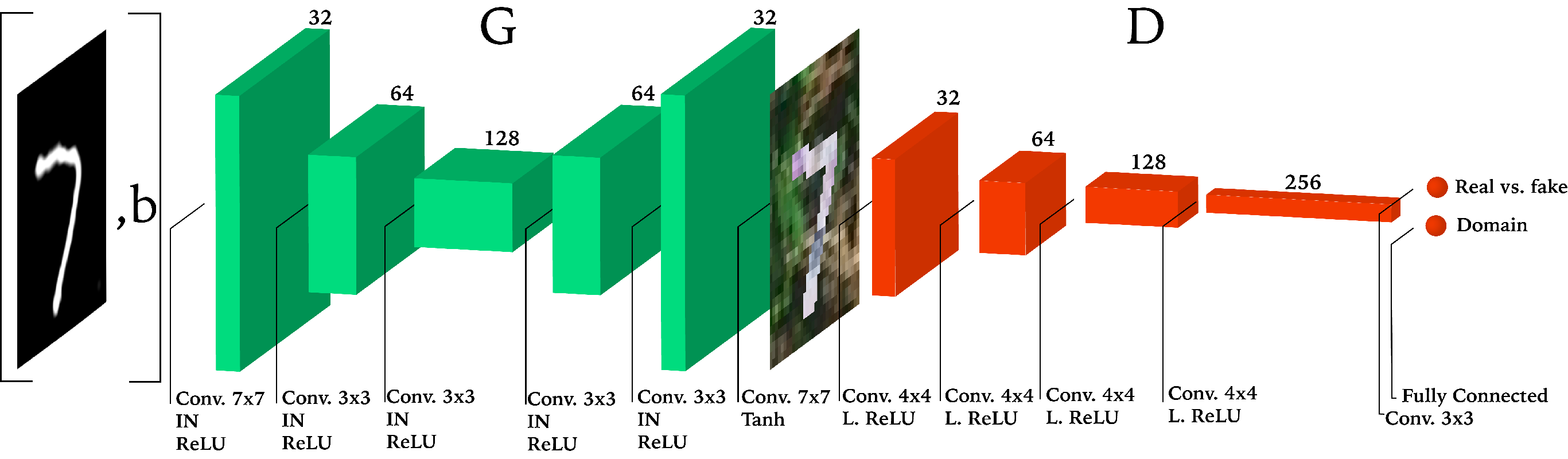} \\
         \includegraphics[height=3.7cm]{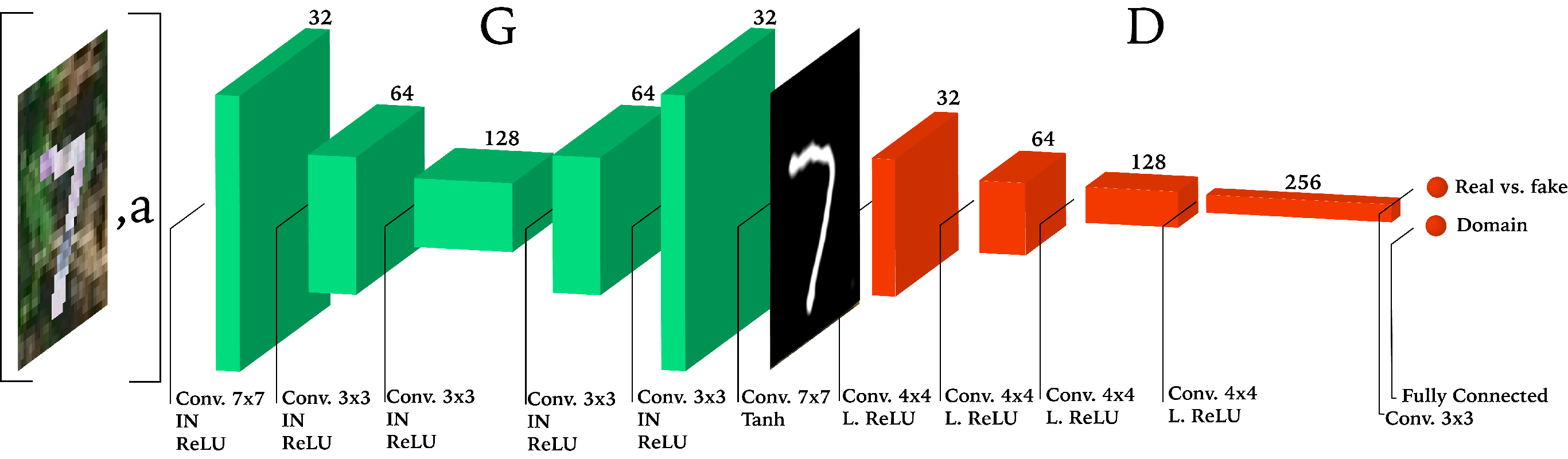}
    \end{tabular}
    \caption{Architecture of the \textit{StarGAN} \cite{stargan} presented in Section~\ref{sec:back-stargan}. \textit{IN} indicates instance normalization and \textit{L. ReLU} leaky rectified linear unit.}
    \label{fig:stargan:adv}
\end{figure*}

Similarly to the \textit{CycleGAN} model explained in Section~\ref{sec:back-cycgan}, the generator $G$ is a FCN, which in our implementation has an initial convolutional block, followed by 2 downsampling blocks, 2 upsampling blocks and a final convolution that maps back to the image domain. Again, as in \textit{CycleGAN}, the discriminator $D$ follows the \textit{PatchGAN}~\cite{transcan} discriminator architecture with 4 downsampling blocks and two ouputs $D_{rf}$ and $D_{dom}$. The network is shown in Figure~\ref{fig:stargan:adv}.

In order to train these modules, three different objectives are defined following similar goals to the ones \textit{CycleGAN}:

\minisection{Translation adversarial loss.} In order to make the generated image-translations look realistic, an adversarial objective is used between $G$ and $D_{rf}$, implemented as the Wasserstein loss with gradient penalty~\cite{wgan-gp}.
\begin{align}
\begin{split}
    \mathcal{L}_{rf}^D =\;&
    \mathbb{E}_{x,\bar{l}}\! \left[D_{rf} (G(x, \bar{l}) )\right]
    - \mathbb{E}_{x}\!\! \left[D_{rf}^t (x)\right] \\
    &\!+ \lambda_{gp}\, \mathbb{E}_{\hat{x}}\!\! \left[(
        \left\lVert \nabla_{\hat{x}} D_{rf} (\hat{x}) \right\rVert_{2}
    \shortminus 1)^2 \right],
    \label{eq:sgan:rfd}
\end{split}
\\[1ex]
    \mathcal{L}_{rf}^G =\;&
    \shortminus \mathbb{E}_{x,\bar{l}}\!\left[D_{rf} (G(x, \bar{l}) )\right],
    \label{eq:sgan:rfg}
\end{align}
where $\hat{x}$ is sampled uniformly along a straight line between a pair of real and generated images and $\lambda_{gp}$ controls the relative importance of the \textit{gradient penalty} component. The setting of generator and discriminator networks to compute these loss components is represented in Figure~\ref{fig:stargan:adv}.

\minisection{Domain classification loss.} The goal of translation is to have the transformed images look from the translated domain, which can also be interpreted as having them classified as belonging to the translation domain. To do so, $D_{dom}$ is trained to classify real images as belonging to their corresponding domain and $G$ is trained to make the translated images correctly classified by $D_{dom}$. This training is performed by minimizing
\begin{align}
    \mathcal{L}_{dom}^{D} =&\;
    \mathbb{E}_{x_a}\! [H (D_{dom} (x_a), a)] 
    + \mathbb{E}_{x_b} [H (D_{dom} (x_b), b)] \ , 
    \label{eq:sgan:domd}
\\[1ex]
\begin{split}
    \mathcal{L}_{dom}^{G} =&\;
    \mathbb{E}_{x_a}\! [H (D_{dom} (G (x_a, b)), b)] \\
    &+ \mathbb{E}_{x_b} [H (D_{dom} (G (x_b, a)), a)] \ , 
    \label{eq:sgan:domg}
\end{split}
\end{align}
where $H$ denotes the \textit{cross-entropy} loss defined as $H(x,y) = -y \cdot \log(x) - (1-y) \cdot \log(1-x)$. The setting of the generator and discriminator networks to compute these loss components is analogous to the translation adversarial component shown in Figure~\ref{fig:stargan:adv}.

\begin{figure}[ht]
    \centering
    \includegraphics[width=0.7\linewidth]{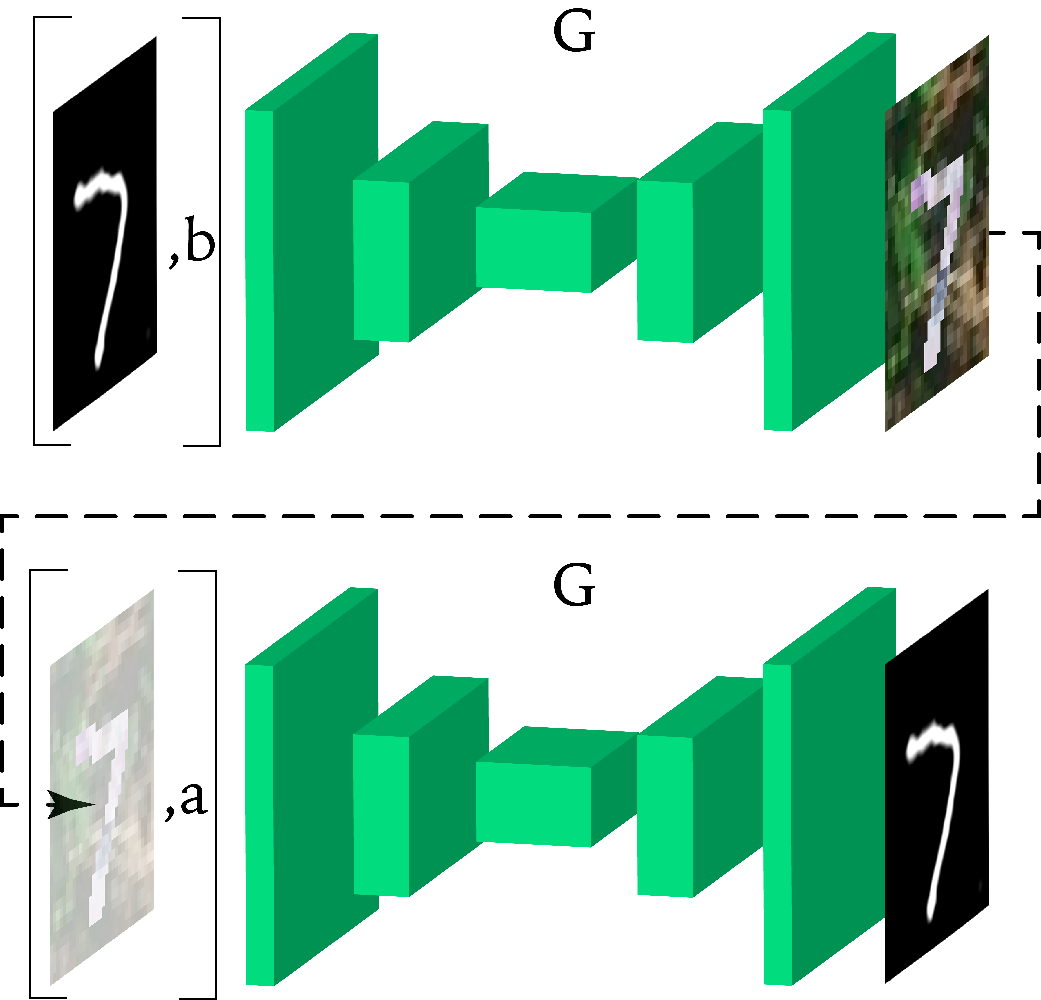}
    \caption{Cycle configuration of the modules in \textit{StarGAN} \cite{stargan} in order to compute $\mathcal{L}_{cyc}$ for domain $a$.}
    \label{fig:stargan:cyc}
\end{figure}

\minisection{Cycle consistency loss.} In order to preserve the content of the input image in the translations, a cycle consistency loss is added to the generator's set of losses, following \textit{CycleGAN}:
\begin{equation}
\begin{split}
    \mathcal{L}_{cyc} =&\;
        \mathbb{E}_{x_a}\!\! \left[ \left \lVert
            G(G(x_a,b), a) - x_a
        \right \rVert_{1} \right]\\ 
        &+ \mathbb{E}_{x_b}\!\! \left[ \left \lVert
            G(G(x_b, a),b) - x_b
        \right \rVert_{1} \right] \label{eq:sgan:cyc} \ ,
\end{split}
\end{equation}
An example of one of the two cycle settings is represented in Figure~\ref{fig:stargan:cyc}.

To wrap up, the model is trained by minimizing $\mathcal{L}^{G}$ for the generator $G$ and $\mathcal{L}^D$ for the discriminator $D$:
\begin{eqnarray}
        \mathcal{L}^{G} \!\!\!\!&=&\!\!\!\! 
        \lambda_{rf} \cdot \mathcal{L}^{G}_{rf}
        + \lambda_{dom} \cdot \mathcal{L}^{G}_{dom}
        + \lambda_{cyc} \cdot \mathcal{L}_{cyc}
        \\[1ex]
        \mathcal{L}^{D} \!\!\!\!&=&\!\!\!\! 
        \lambda_{rf} \cdot \mathcal{L}^{D}_{rf}
        + \lambda_{dom} \cdot \mathcal{L}^{D}_{dom}
\end{eqnarray}
where $\lambda_{rf}$, $\lambda_{dom}$ and $\lambda_{cyc}$ control the relative importance of the three objectives.

\section{Method}
\label{sec:method}
In the setting of unsupervised domain adaptation, we assume access to a  dataset $\mathcal{D}_s=\{(x_s^{(i)}, y_s^{(i)})\}_{i=1}^{N_s}$ drawn from a source domain distribution, where in the case of image segmentation, $x$ denote images and $y$ ground-truth segmentation masks. At the same time, we assume access to another dataset $\mathcal{D}_t=\{(x_t^{(n)})\}_{n=1}^{N_t}$ sampled from a target domain distribution. Note that in $\mathcal{D}_t$ the ground-truth segmentation masks are not provided. With these two datasets the goal is to learn a function $f: x \rightarrow y$ that correctly predicts both $y_s$ and $y_t$ given $x_s$ and $x_t$, respectively and generalizes properly to unseen source and target samples.

\begin{figure*}[ht]
    \centering
    \includegraphics[height=7.5cm]{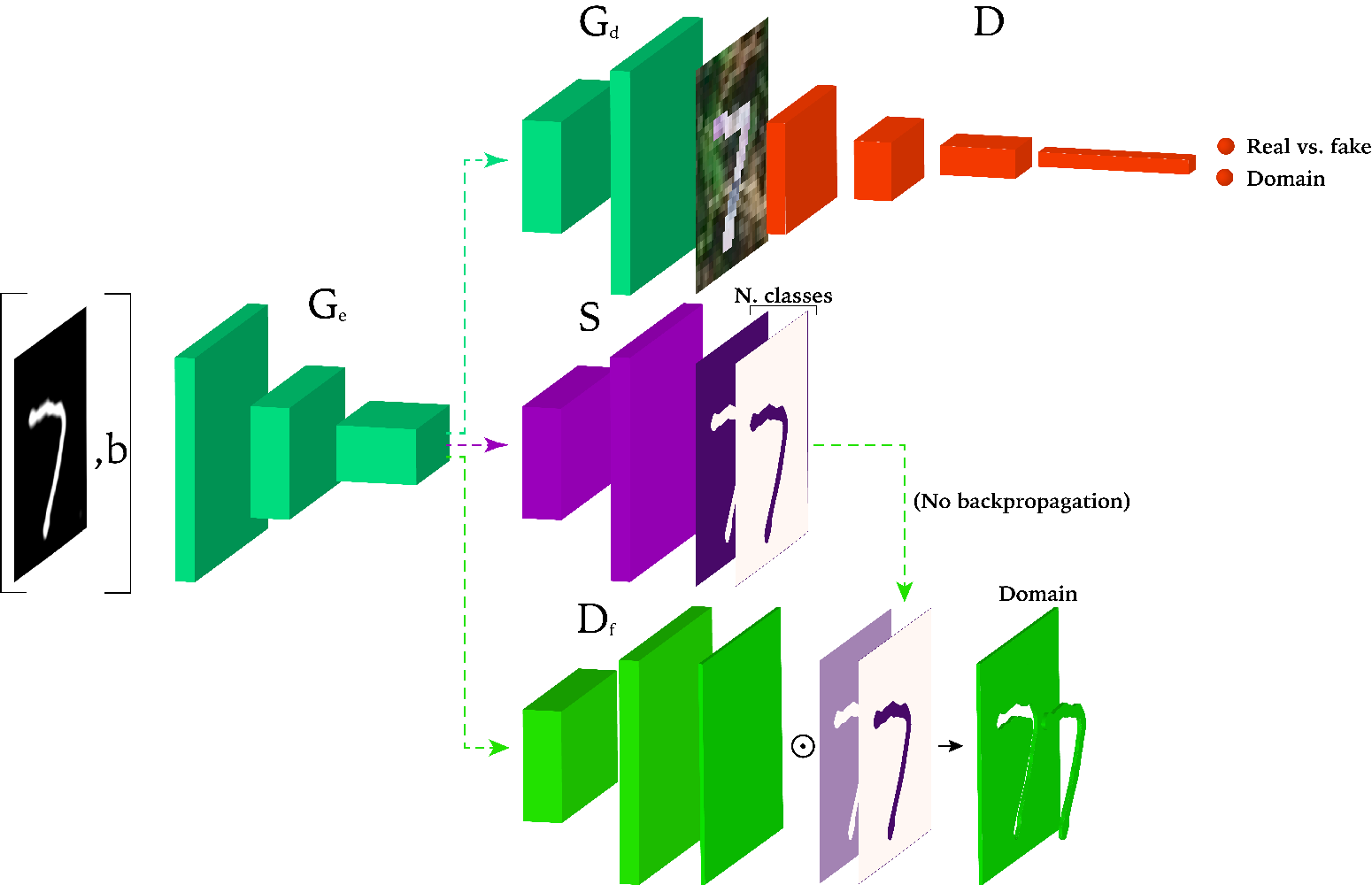}
    \caption{Complete view of the model, with the connections between the different modules: the \textit{StarGAN} generator $G$ (decomposed in an encoder $G_e$ and a decoder $G_d$), the \textit{StarGAN} discriminator $D$, the segmenter $S$ and the feature discriminator $D_f$.}
    \label{fig:ourmodel:all}
\end{figure*}

In order to tackle the unsupervised domain adaptation problem, we introduce a model based on an image translation backbone that we use to translate images $x_s$ and $x_t$ to their opposite domain. Then, a segmenter network is connected to the bottleneck features of the image translator to perform our end task (semantic segmentation). The bottleneck features of the image translator are also connected to an additional module, which performs a class specific feature matching. The rationale behind this feature matching module is to ensure that features belonging to the same class come from the same distribution. An overview of the model is depicted on Figure~\ref{fig:ourmodel:all}: note that $G_e$ and $G_d$ depict the image translation network, followed by a discriminator $D$; $S$ refers to the segmenter; and $D_f$ denotes the class-aware feature matching module. The rest of the section will be devoted to providing details on each component of the proposed model.

\minisection{Image translation backbone}. As mentioned in Section~\ref{sec:sota-domadapt}, many domain adaptation architectures use image translation or image reconstruction as an auxiliary task to obtain more general features, in order to boost the classification or segmentation performance on the target domain. In our model, we use image translation as auxiliary task and we rely on the state-of-the-art \textit{StarGAN} \cite{stargan} model as the backbone module of the proposed network.

As already explained in Section~\ref{sec:back-stargan}, \textit{StarGAN} is able to perform unpaired image translations between a set of different image domains, so we are able to exploit it to translate our images from $\mathcal{D}_s$ (source domain) to $\mathcal{D}_t$ (target domain) and the other way around. The main advantage of using \textit{StarGAN} architecture is that it allows us to have domain specific parameters in a single network, drastically reducing the number of parameters required to perform image translation w.r.t. related approaches in the literature such as \cite{domsep,labelefficient,dirtt,adda,monocular-depth}.

Therefore, our model's base is composed of both \textit{StarGAN} generator $G$ and discriminator $D$ networks. These networks are trained with the \textit{StarGAN} objectives introduced in Section~\ref{sec:back-stargan}. Using $x_s$ and $x_t$ as input images, we define:
\begin{itemize}
    \item $\mathcal{L}_{rf}^{G}$ and $\mathcal{L}_{rf}^{D}$ (Eqs. \eqref{eq:sgan:rfg} and \eqref{eq:sgan:rfd}) to encourage the model to generate realistic looking translations.
    \item $\mathcal{L}_{dom}^{G}$ and $\mathcal{L}_{dom}^{D}$ (Eqs. \eqref{eq:sgan:domg} and \eqref{eq:sgan:domd}) to make the translated images look like drawn from their corresponding translation domain.
    \item $\mathcal{L}_{cyc}$ (Eq. \eqref{eq:sgan:cyc}) to encourage the network to preserve the content of the images while translating them.
\end{itemize}

\minisection{Segmenter}. To perform the segmentation task, we connect a segmenter decoder to the bottleneck of \textit{StarGAN}. To do so, we divide the generator $G$ into the encoder composed of the downsampling path of $G$ and, denoted from now on as $G_e$, and the decoder composed of the upsampling path of $G$ and referred to as $G_d$. This way, we connect a segmentation decoder $S$ to $G_e$, and feed both $G_d$ and $S$ with the output of $G_e$. Note that the segmentation decoder $S$ is built following the upsampling path of the segmenter defined in Section~\ref{sec:back-fcnss}. The above-described connection between $G$ and $S$ is depicted in Figure~\ref{fig:ourmodel:all}.

The segmentation decoder $S$ is trained by minimizing the \textit{soft IOU loss} defined in Eq. \eqref{eq:segm:siou} between $y_s$ and $\hat{y}=S(G_e(x_s))$ for the binary segmentation case. Recall that in a multi-class segmentation setting, the loss used to train the segmenter would be the \textit{cross-entropy} loss defined in Eq. \eqref{eq:segm:ce}. It is worth mentioning that, while minimizing the segmentation loss, gradients are also backpropagated through $G_e$.

\minisection{Feature matching.} With the previous components, our model is capable of segmenting images by learning from the source dataset $\mathcal{D}_s$ while obtaining signal from the target images in $\mathcal{D}_t$ through the image translation task. However, this does not ensure that the segmenter will be able to segment properly the target images $x_t$ with the features obtained from the encoder $G_e$. In order to endow the network with this ability, a common approach in the domain adaptation literature is to perform a matching between the encoded features from images from both domains. This matching is commonly performed in a global way, often ignoring the distribution of classes in each domain and thus, in the case of semantic segmentation, failing to consider the domain-specific shapes and sizes of the elements. In this thesis, we introduce a class conditional adversarial feature matching with the aim of easing the above-mentioned problem.

\begin{figure}[ht]
    \centering
    \includegraphics[width=0.8\linewidth]{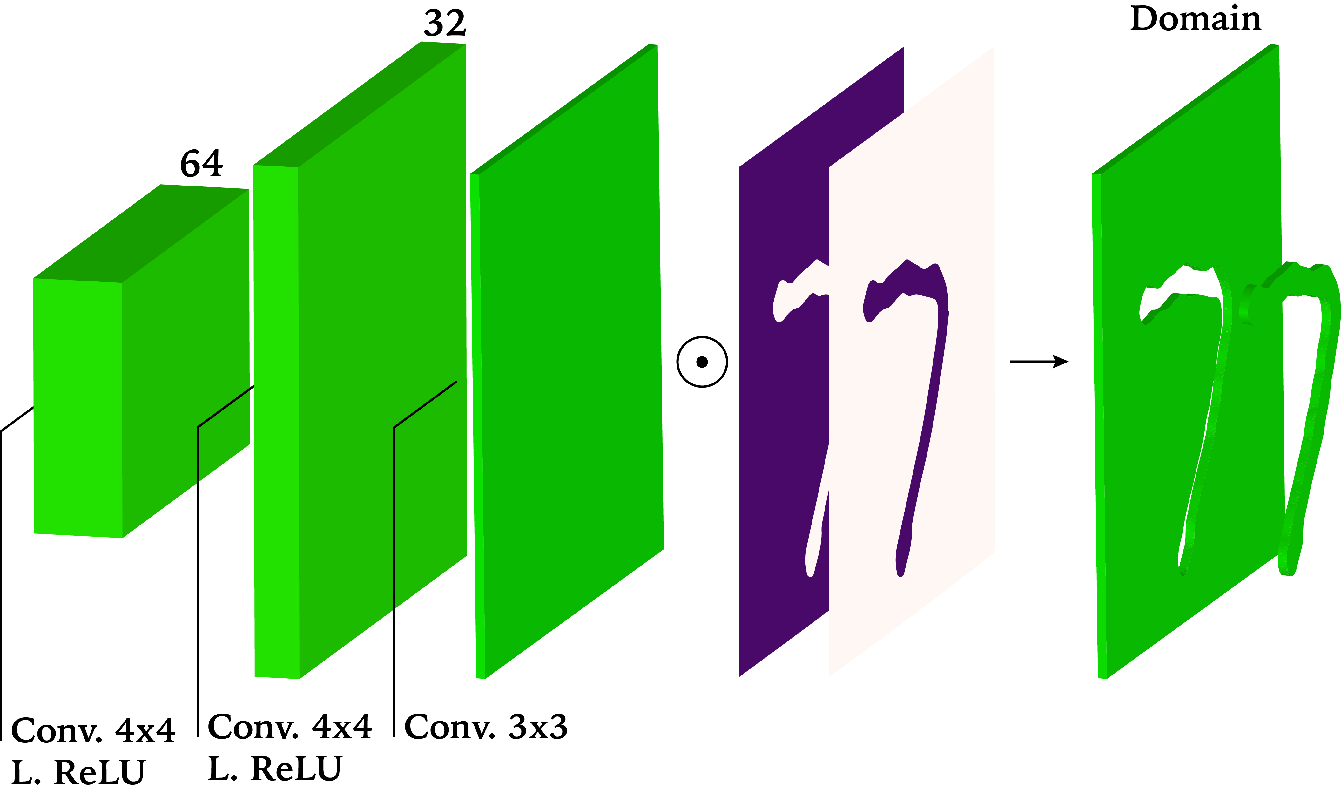}
    \caption{Structure of the feature discriminator $D_f$ of our model, in addition to a representation of the post processing of its output. $D_f$ upsamples the features to spatial resolution predicting whether each pixel in the image space belongs to the source domain or the target domain. Then the prediction of the segmentation masks are applied to this prediction in order to obtain the probabilities separated per class. \textit{L. ReLU} indicates leaky rectified linear unit.}
    \label{fig:ourmodel:df}
\end{figure}

We define a discriminator network $D_f$ that upsamples the features $h_s = G_e(x_s,s)$ and $h_t = G_e(x_t,s)$ to their original spatial resolution, with a single channel output. $D_f$ is implemented by two upsampling convolution and a final convolutional layer and its role is to predict, whether each pixel in the image space belongs to the source domain or the target domain.  Then, the segmentation prediction from $S(h_{\bullet})$ is applied to the output of the discriminator $D_f$ with the goal of obtaining per class probabilities of features coming from source and target domains, respectively. These per class probabilities are computed as follows:
\begin{equation}
    \frac {\sum_{p} \hat{y}_\bullet^{c,p} \cdot D_f(h_\bullet)^{c,p}}
        {\sum_{p} \hat{y}_\bullet^{c,p}} \ ,
\end{equation}
where $D_f(h_\bullet)^{c,p}$ denotes the value of the output of $D_f(h_\bullet)$ for a certain class $c$ and pixel $p$. Analogously, we denote as $\hat{y}_\bullet^{c,p}$ the value of the segmentation prediction $\hat{y}_s=S(h_s)$ for the same given pixel and class. Note that this formulation are independent of the domain and will be used with both source and target features and segmentations. A representation of the output post-processing and the details of $D_f$ are shown in Figure~\ref{fig:ourmodel:df}. 

Thus, the objective of $D_f$ is to confidently distinguish the original domain of the features per class, whereas the objective of $G_e$ is to effectively fool $D_f$ by trying to match the per class distribution of features. Following this objective, we define the loss to be optimized by $D_f$, denoted $\mathcal{L}_{dom}^{D_f}$, as:
\begin{equation}
\begin{split}
    %\mathcal{L}_{dom}^{D_f} =& \enskip 
    \mathbb{E}_{h_t,c}\!\! \left[
    \frac
        {\sum_{p} \hat{y}_t^{c,p} \!\cdot\! D_f(h_t)^{c,p}}
        {\sum_{p} \hat{y}_t^{c,p}} 
    \right] 
    -\mathbb{E}_{h_s,c}\!\! \left[
    \frac
        {\sum_{p} \hat{y}_s^{c,p} \!\cdot\! D_f(h_s)^{c,p}}
        {\sum_{p} \hat{y}_s^{c,p}} 
    \right] \\[1ex]
    +\lambda_{gp} \cdot \mathbb{E}_{\bar{h},c}\!\! \left[
    \left( \left\lVert \nabla_{\bar{h}} \left(
    \frac
        {\sum_{p} \bar{y}^{c,p} \!\cdot\! D_f(\bar{h})^{c,p}}
        {\sum_{p} \bar{y}^{c,p}} 
     \right) \right\rVert_2 \shortminus 1 \right)^2 \right],
\end{split}
\end{equation}
Note that this loss follows the \textit{Wasserstein with gradient penalty} \cite{wgan-gp} formulation, where $\bar{h}$ and $\bar{y}$ are sampled uniformly along a straight line between a pair of source and target features, and segmentations, respectively. Following previous notation, $\lambda_{gp}$ represents the hyper-parameter controlling the relative importance of the \textit{gradient penalty} component.

At the same time, $G_e$ is trained by minimizing the following objective function $\mathcal{L}_{dom}^{G_e}$:
\begin{equation}
    %\mathcal{L}_{dom}^{G_e} =  
    \mathbb{E}_{h_s,c}\!\! \left[
    \frac
        {\sum_{p} \hat{y}_s^{c,p} \!\cdot\! D_f(h_s)^{c,p}}
        {\sum_{p} \hat{y}_s^{c,p}} 
    \right] 
    -\mathbb{E}_{h_t,c}\!\! \left[
    \frac
        {\sum_{p} \hat{y}_t^{c,p} \!\cdot\! D_f(h_t)^{c,p}}
        {\sum_{p} \hat{y}_t^{c,p}} 
    \right].
\end{equation}
Note that the gradients obtained from the segmentation predictions should neither be backpropagated through $S$ nor $G_e$, as the segmentations are only used for the aggredation of the values per class.
\bigskip

Overall, the model is trained by minimizing the joint objectives of each module. More specifically, $G$ and $S$ are trained by minimizing
\begin{equation}
\begin{split}
        \mathcal{L}^{G,S} =\;&
        \lambda_{rf} \!\cdot\! \mathcal{L}^{G}_{rf}
        + \lambda_{dom} \!\cdot\! \mathcal{L}^{G}_{dom}
        + \lambda_{cyc} \!\cdot\! \mathcal{L}_{cyc}\\
        &\!\!+ \lambda_{segm} \!\cdot\! \mathcal{L}_{segm}
        + \lambda_{dom}^f \!\cdot\! \mathcal{L}^{G_e}_{dom} \ ,
\end{split}
\end{equation}
where $\lambda_{rf}$, $\lambda_{dom}$, $\lambda_{cyc}$, $\lambda_{segm}$ and $\lambda^f_{dom}$ control the relative importance of the five objectives; emphasizing that gradients obtained for segmentation predictions in $\mathcal{L}^{G_e}_{dom}$ are not backpropagated.

Furthermore, the discriminator $D$ from the \textit{StarGAN} module is trained by minimizing
\begin{equation}
        \mathcal{L}^{D} =
        \lambda_{rf} \!\cdot\! \mathcal{L}^{D}_{rf}
        + \lambda_{dom} \!\cdot\! \mathcal{L}^{D}_{dom} \ ,
\end{equation}
where $\lambda_{rf}$ and $\lambda_{dom}$ are the same values used in $\mathcal{L}^{G,S}$.

Finally, we train the feature discriminator $D_f$ by minimizing the objective
\begin{equation}
        \mathcal{L}^{D_f} = \mathcal{L}^{D_f}_{dom}.
\end{equation}

\section{Experiments}
\label{sec:exp}
In this section, we will introduce and detail the experiments we have performed to measure and prove the improvements of our model with respect to some baselines and to study the relative importance of the improvement with a model ablation.

\subsection{Datasets}
The datasets used for the experiments of this thesis are selected and designed to execute a proof of concept of the contributions. Given this goal, the datasets have been prepared to simulate some features that we encounter in real scenarios, namely the differences in appearance between different domains, as well as the per class pixel distributions. This has allowed us for a more dynamic exploration and analysis of different alternatives, which will be essential to tackle bigger and more complex datasets with a higher degree of certainty with respect to the proposed model, as future work.

\begin{figure*}[ht]
    \centering
    \begin{tabular}{rl}
    MNIST       & \includegraphics[height=1cm]{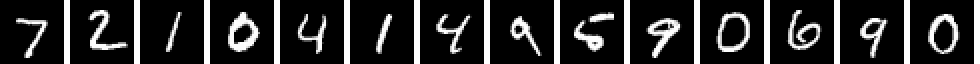} \\
    MNIST-M     & \includegraphics[height=1cm]{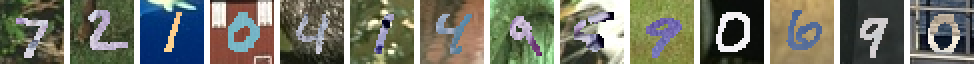} \\
    MNIST-thin  & \includegraphics[height=1cm]{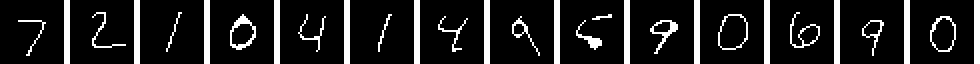} 
    \end{tabular}
    \caption{Dataset samples from MNIST, MNIST-M and MNIST-thin. Note the changes in texture, color and brightness of MNIST-M vs MNIST as well as the changes in per class pixel distributions of MNIST vs MNIST-thin.}
    \label{fig:dset-samples}
\end{figure*}

% \footnote{http://yann.lecun.com/exdb/mnist/}
\minisection{MNIST.} The first dataset used for the experiments is the MNIST handwritten digit dataset from LeCun et al. \cite{mnist}. It contains $60,000$ training samples and $10,000$ test samples. Following common practice, we split the training set in $50,000$ training samples and $10,000$ validation samples ($\%16.\hat{6}$ of the original training set). Some samples from the dataset are shown in Figure~\ref{fig:dset-samples}a.

\minisection{MNIST-M\footnote{http://yaroslav.ganin.net/}.} We use the modification of MNIST presented in \cite{unsudomadapt}, which adds texture and color to the original version of MNIST, using random patches from BSDS500 dataset \cite{BSDS500} and inverting the color in the pixels belonging to the digit. Particulary, considering $I^M$ an MNIST image and $I^B$ a BSDS500 random patch (both in range $[0,1]$), the MNIST-M image is obtained as $I=|I^B-I^M|$. A few examples from the MNIST-M dataset are shown in Figure~\ref{fig:dset-samples}b. Note the varied textures and colors w.r.t. the original dataset. 

The interesting characteristic about this dataset is that, when using it as a target domain in the domain adaptation problem, given a source domain as MNIST in simple black and white, it simulates the knowledge transfer from simulated environments to more complex real ones. It emulates the domain shift with colors, textures and different lightning conditions that could be found in this situations.

\minisection{MNIST-thin.} To obtain masks with different per class pixel distributions with respect to MNIST and MNIST-M, we generate another modification of the MNIST dataset, which is based on eroding the original digits. The process to create this new dataset is the following: we resize the MNIST images to $64\times64$; then we apply an erosion of $4$ pixel radius disk and; finally, we add the skeleton of the original digit \cite{skeleton}. Some examples resulting from this process are shown in Figure~\ref{fig:dset-samples}c.

This modification allows us to simulate different class distributions across datasets, obtaining a $\%5.18$ of mean foreground (digit) area per image in this dataset, compared to the original $\%14.64$ of the MNIST and MNIST-M datasets. 

\subsection{Baselines}
\label{sec:exp:base}
In this section, we present the baselines we have tried to overtake with our model. We train these baselines with MNIST-thin as source domain and MNIST-M as target domain, highlighting the difference in appearance and per class distribution among the two domains, and providing an insightful proof of concept. The choice of this dataset pair is motivated by the challenges it poses, namely training on simple black and white images and generalizing to a complex domain with textures and colors, as well as the challenge of dealing with different foreground/background ratios. 

\minisection{Single FCN.} The most simple approach to overtake in the unsupervised domain adaptation problem is the one where (following the notation of Section~\ref{sec:method}) the model is trained using the source domain dataset $\mathcal{D}_s$, exclusively, and tested on the target domain out-of-the-box.

In our case, we train the FCN segmenter from Section~\ref{sec:back-fcnss} only using  $\mathcal{D}_s$. In particular, we train the network on MNIST-thin and we test it on MNIST-M images. As it is a binary segmentation problem, we train the model by minimizing the \textit{soft IOU loss} of Equation \eqref{eq:segm:siou}. The results of this setting are aimed to provide a lower bound on the performance we can achieve, given that the baseline does not include any adaptation step and due to the existing differences between the two domains. 

\minisection{StarGAN with segmenter.} In order to exploit both $\mathcal{D}_s$ and $\mathcal{D}_t$ simultaneously, one could resort to training with additional auxiliary losses, e.g. for image reconstruction or image translation between domains. In this baseline, we focus on the later auxiliary task, since it has proven to be successful in the literature \cite{domadaptgener}. Note that this baseline is analogous to our model, described in Section~\ref{sec:method}, without the feature matching objectives and the class-aware discriminator $D_f$. The design of the model can be inferred by ignoring the $D_f$ module in Figure~\ref{fig:ourmodel:all}.

In essence, the model is trained with the \textit{StarGAN} objectives from Section~\ref{sec:back-stargan}, together with the \textit{soft IOU loss} from Equation \eqref{eq:segm:siou}. We train the generator $G$ and the segmenter $S$ by minimizing the objective
\begin{equation}
        \mathcal{L}^{G,S} =
        \lambda_{rf} \!\cdot\! \mathcal{L}^{G}_{rf}
        + \lambda_{dom} \!\cdot\! \mathcal{L}^{G}_{dom}
        + \lambda_{cyc} \!\cdot\! \mathcal{L}_{cyc}
        + \lambda_{segm} \!\cdot\! \mathcal{L}_{segm} \ ,
\end{equation}
where $\lambda_{rf}$, $\lambda_{dom}$, $\lambda_{cyc}$ and $\lambda_{segm}$ control the relative importance of the five objectives. Following the same line, we train the discriminator $D$ by minimizing
\begin{equation}
        \mathcal{L}^{D} =
        \lambda_{rf} \!\cdot\! \mathcal{L}^{D}_{rf}
        + \lambda_{dom} \!\cdot\! \mathcal{L}^{D}_{dom} \ ,
\end{equation}
where $\lambda_{rf}$ and $\lambda_{dom}$ are the same parameters defined for $\mathcal{L}^{G,S}$.

Note that this model does not have any domain adaptation component, but should still obtain more general features than the previous single FCN baseline. For this reason, we should expect better results than the single FCN baseline, providing a more competitive lower bound.

\subsection{Model ablation}
\label{sec:exp:abl}
By means of this study, we aim to assess the impact of the contributions of this thesis: the convenience of using a single image translation network with domain specific parameters and the advantage of the proposed class-conditional feature matching discriminator. Special attention will be devoted to considering alternative designs of the latter.

\subsubsection{\textbf{Image-translation architectures}}
\label{sec:exp:abl:imtrans}
In the setting of our problem, it is important to have a good image translation base model in order to obtain better features to be exploited by images from the target domain. Nevertheless, considering image translation as an auxiliary task, it is also important not to waste excessive capacity in this component (note that the decoder part of the image translator will be thrown away at test time). For this particular reason and to highlight the benefit of having domain specific encoder parameters in a single image translation network, we define this experiment to compare and contrast the results achieved by \textit{CycleGAN} (Section~\ref{sec:back-cycgan}) w.r.t. the ones achieved by \textit{StarGAN} (Section~\ref{sec:back-stargan}).

Our aim is to determine whether, despite the significant reduction of number of parameters in \textit{StarGAN} and given a similar training strategy, the resulting translations are qualitatively comparable.

\subsubsection{\textbf{Class conditional discriminator}}
\label{sec:exp:abl:condis}
In order to measure the benefits of conditioning the feature adversarial matching per classes (as described in Section~\ref{sec:method}), we define additional experiments using different modifications of $D_f$. A natural first step, is to compare our class conditional discriminator to an unconditional one. This allows us to asses the improvement achieved by our per class feature matching with respect to an often used global feature matching. Our hypothesis is that the conditioning should help in the matching due to the access to a more local discrimination. Furthermore, we define another version of a class conditioned discriminator based on the conditioning used in \textit{StarGAN}, where the labels are given at the input. Conditioning the discriminator at input level is a natural alternative to the proposed framework, which is potentially at higher stake of making the task too easy for the discriminator. More details on these alternative discriminator designs are provided in the remainder of this section. 

\begin{figure*}[t]
    \centering
    \subfloat[Structure of the input conditional feature discriminator.]{
        \includegraphics[width=0.55\textwidth]{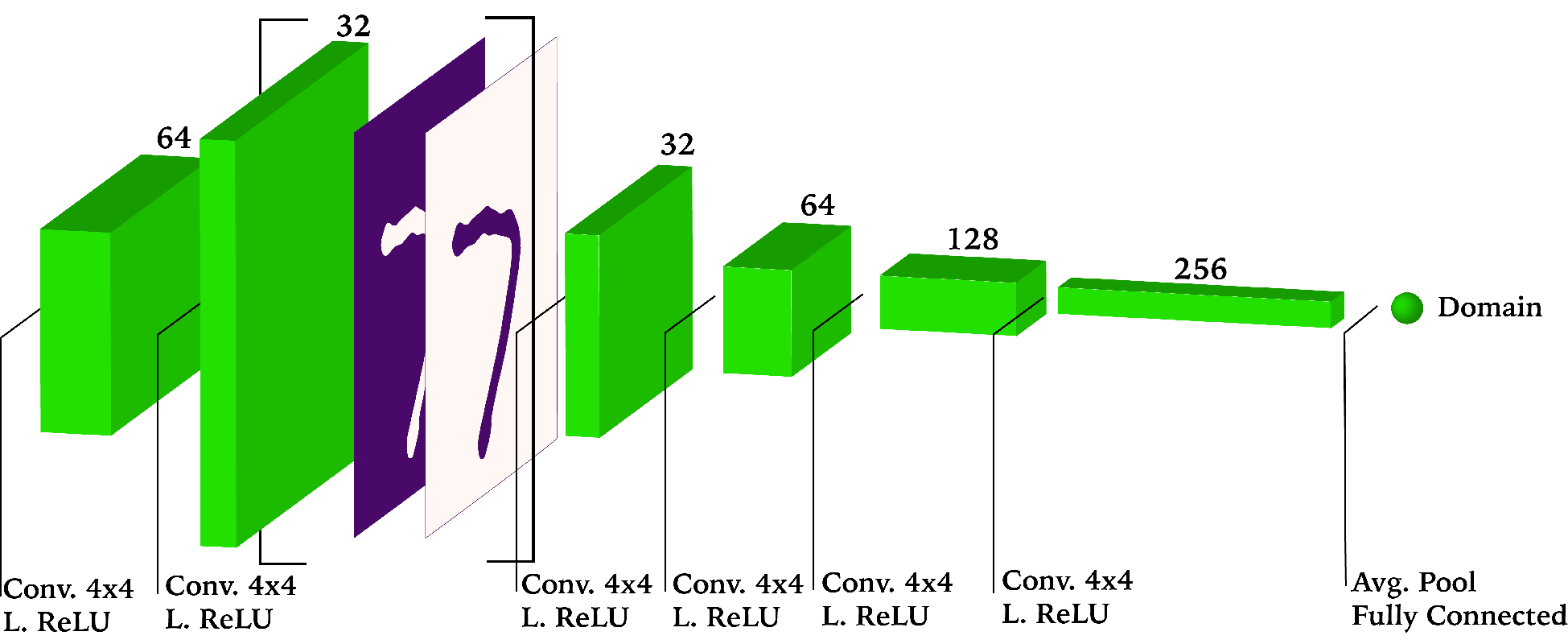}
        \label{fig:df:input}
    }
    \rulesep
    \subfloat[Structure of the unconditional feature discriminator.]{
        \includegraphics[width=0.40\textwidth]{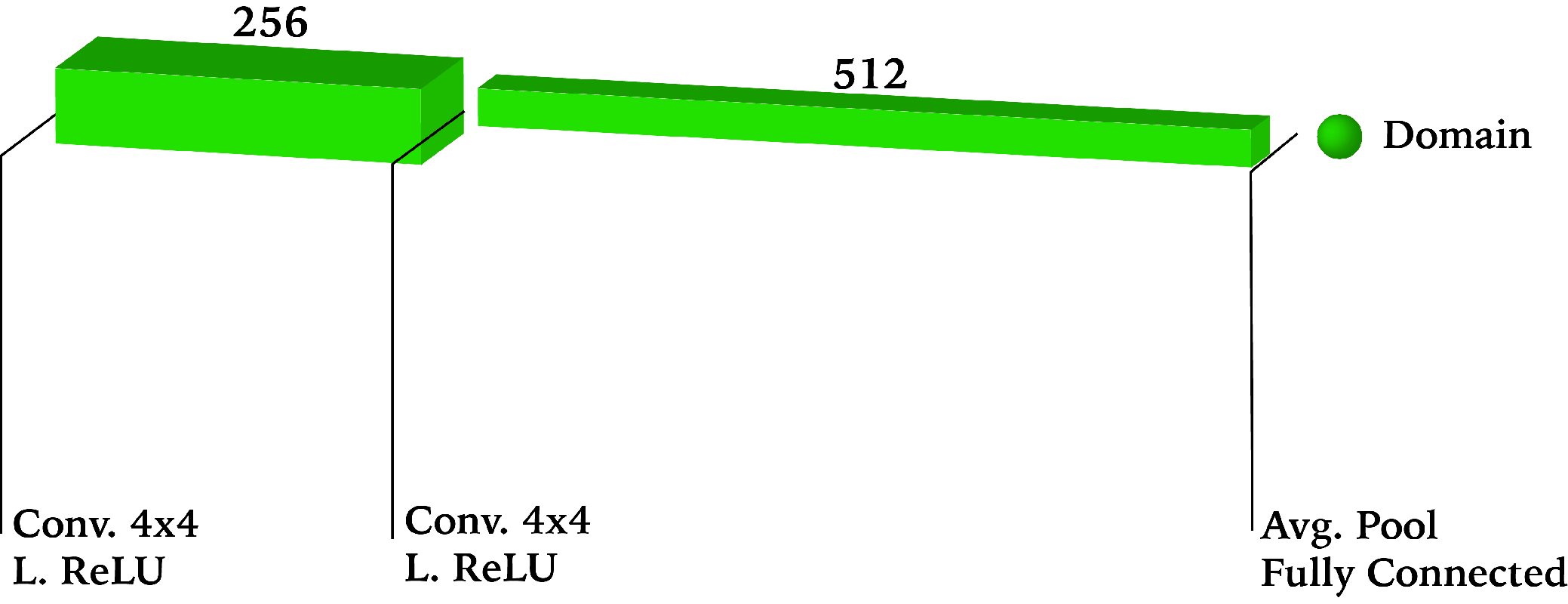}
        \label{fig:df:uncond}
    }
    \caption{Alternative feature discriminators introduced in Section~\ref{sec:exp:abl:condis}.}
    \label{fig:df}
\end{figure*}

\minisection{Unconditional discriminator.} In this case, the discriminator network $D_f$ tries to distinguish source and target domains from global features, with no class conditioning. The discriminator network is implemented by 2 downsampling convolutions, an average pooling and a fully connected classification layer. The network is depicted in Figure~\ref{fig:df:uncond}. Note that the output of this discriminator has now spatial resolution, meaning it only outputs a single score (source vs target) for its input features. The rest of the modules remain the same as our model from Section~\ref{sec:method}. 

Again, the adversarial matching is performed by means of the \textit{Wasserstein adversarial loss with gradient penalty} \cite{wgan-gp}, redefining the following objectives w.r.t. the porposed model:
\begin{align}
\begin{split}
    \mathcal{L}_{dom}^{D^f} =\;&
    \mathbb{E}_{h_t} \left[D^f (h_t)\right]
    -\mathbb{E}_{h_s} \left[D^f (h_s) )\right] \nonumber \\
    &\!+\lambda_{gp} \!\cdot\! \mathbb{E}_{\bar{h}} \left[
        \left(\left\lVert
            \nabla_{\bar{h}} D^f (\bar{h}) 
        \right\rVert_2 \shortminus 1\right)^2
        \right] 
\end{split}
\\[1ex]
    \mathcal{L}^G =\;&
    \mathbb{E}_{h_s} [D^f (h_s)]
    -\mathbb{E}_{h_t} [D^f (h_t)]
\end{align}

\minisection{Input conditional discriminator.} With this discriminator $D_f$, we aim to perform the matching of the features per class, as in our model, but providing the segmentation prediction information at input level instead. This is achieved by concatenating the segmentation prediction to the features extracted by the encoder, following the same idea as \textit{StarGAN} with the translation labels. Note that if the encoder contains downsampling operations (like in our case), the segmentation predictions do not match the resolution of the extracted features. To overcome this situation, the network is designed with an upsampling path with 2 transposed convolutional blocks to recover the segmentation resolution. The upsampling path is applied to the features extracted by the encoder prior to concatenation with the segmentation prediction. After concatenation, the data is processed by 4 downsampling convolutional blocks followed by a fully connected layer. The architecture of the discriminator module is shown in Figure~\ref{fig:df:input}. The rest of the modules remain untouched from our model.

Once again, the adversarial matching is performed by means of the \textit{Wasserstein adversarial loss with gradient penalty} \cite{wgan-gp}, redefining the following objectives from our model:
\begin{align}
\begin{split}
    \mathcal{L}_{dom}^{D^f} =\;&
    \mathbb{E}_{h_t} \left[D^f (h_t,\hat{y}_t)\right]
    -\mathbb{E}_{h_s} \left[D^f (h_s,\hat{y}_s) )\right] \\
    &\!+\lambda_{gp} \!\cdot\! \mathbb{E}_{\bar{h}} \left[
        \left(\left\lVert
            \nabla_{\bar{h}} D^f (\bar{h},\bar{y}) 
        \right\rVert_2 \shortminus 1\right)^2
        \right] 
\end{split}
\\[1ex]
    \mathcal{L}_{dom}^G =\;&
    \mathbb{E}_{h_s} \left[D^f (h_s,\hat{y}_s)\right]
    -\mathbb{E}_{h_t} \left[D^f (h_t,\hat{y}_t)\right]
\end{align}

\section{Results}
\label{sec:res}
In this section, we will present and analyze the results obtained in the previously defined experiments. First, we will show qualitative results comparing \textit{CycleGAN} and \textit{StarGAN} as outlined in Section~\ref{sec:exp:abl:imtrans}. Then,  we will report quantitative results comparing our model to the baselines and ablated models introduced in Sections~\ref{sec:exp:base} and~\ref{sec:exp:abl}, respectively. 

\subsection{Image translation results}
The goal of this experiment is to assess the image translation quality of \textit{StarGAN} vs \textit{CycleGAN} and argue the choice of the image translation backbone in the proposed unsupervised domain adaptation model. To do so, we train both \textit{StarGAN} and \textit{CycleGAN} models to perform image translation between MNIST and MNIST-M images.  In order to compare both models, we provide a small set of qualitative examples in Figure~\ref{fig:translations} (additional examples are included in Appendix~\ref{apx:extranslations}). Figure~\ref{fig:translations} depicts sample translations from MNIST to MNIST-M (left) and sample translations from MNIST-M to MNIST (right) for both models. By means of these translation samples, we observe that \textit{StarGAN} results are qualitatively comparable to those of \textit{CycleGAN}. It is worth highlighting the variety of colors and textures in MNIST-M translations achieved by the trained \textit{StarGAN}, as well as the rather sharp MNIST translations. \textit{CycleGAN} exhibits slightly more uniform colors and textures, and less sharp digit translations. These details can be better perceived in the samples shown in Appendix~\ref{apx:extranslations}.

\begin{figure*}[ht]
    \centering
    \begin{tabular}{rll}
    Image               & \includegraphics[height=1cm]{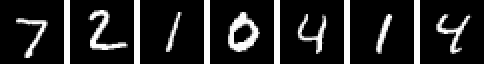}       & \includegraphics[height=1cm]{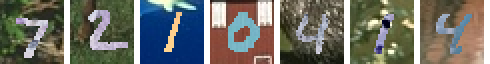} \\
    \textit{CycleGAN}   & \includegraphics[height=1cm]{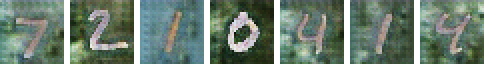} & \includegraphics[height=1cm]{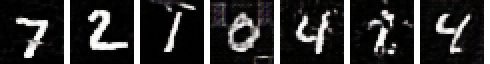} \\
    \textit{StarGAN}    & \includegraphics[height=1cm]{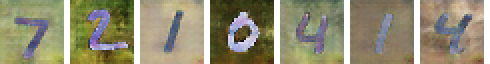}     & \includegraphics[height=1cm]{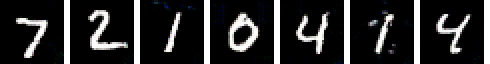} 
    \end{tabular}
    \caption{Example of translations obtained from \textit{CycleGAN} and \textit{StarGAN} between MNIST and MNIST-M datasets. Left: translations from MNIST to MNIST-M. Right: translations from MNIST-M to MNIS.}
    \label{fig:translations}
\end{figure*}

Along with the translation samples, we detail the number of parameters required by a standard implementation of both models, see Table~\ref{tab:nparams}. As shown in the table, \textit{CycleGAN} and \textit{StarGAN} use similar architectures for their generators and discriminators. Recall that \textit{CycleGAN} requires two generators and two discriminators, whereas \textit{StarGAN} only requires a single generator and a single discriminator. Therefore, \textit{StarGAN} exhibits a reduction of roughly $50\%$ in number of parameters when compared to \textit{CycleGAN}. The extra parameters in the \textit{StarGAN} generator and discriminator networks (w.r.t. the \textit{CycleGAN} single domain networks) are caused by the domain input conditioning of the generator and the domain classification output of the discriminator. This small difference is negligible compared to the overall drop in number of parameters, when scaling the networks to higher capacities.
\begin{table}[b]
    \centering
    \label{tab:nparams}
    \begin{tabular}{@{}lcc@{}}
    \toprule
                                       & \textbf{Generator \# parameters} & \textbf{Discriminator \# parameters} \\ \midrule
    \multirow{2}{*}{\textbf{CycleGAN}} & $G_{ab}=194,051$                 & $D_a=694,241$                        \\
                                       & $G_{ba}=194,051$                 & $D_b=694,241$                        \\ [0.2cm]
    \textbf{StarGAN}                   & $197,504$                        & $694,496$                            \\ \bottomrule
    \end{tabular}
    \vspace{0.2cm}
    \caption{Comparison of number of parameters of state-of-the-art image translation models.}
\end{table}
\vspace*{-5mm}
\subsection{Unsupervised domain adaptation results}
In this subsection, we provide both qualitative and quantitative results of the different baselines and model ablations from Section~\ref{sec:exp:abl}, together with the results from our  model. All models are trained to perform unsupervised domain adaption from MNIST-thin to MNIST-M, i.e. only MNIST-thin segmentation masks are used for training. Recall that the goal the experiments is to assess how the proposed model performs when adapting between domains exhibiting different appearance as well as class distributions. Results are reported in terms of per class Intersection over Union (IoU) as well as mean IoU (mIoU) on both source and target domains. IoU is a common metric used in semantic segmentation and is computed as
\begin{equation}
    \frac
        {TP}
        {TP+FN+FP} \ ,
\end{equation}
where TP, FN and FP denote the true positive, false negative and false positive values between a ground-truth and a prediction mask for a certain specific class. Note that given the binary segmentation masks, the maximum desired value is $1$ and the minimum is $0$ for each class and that the mIoU is computed as the mean over the IoU values from each class.

\begin{figure*}[t!]
    \centering
    \begin{tabular}{rl}
Image          & \includegraphics[height=1cm]{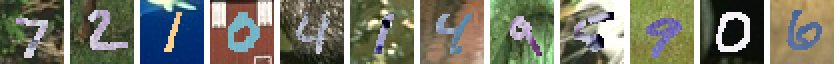} \\
\textit{FCN Segmenter}           & \includegraphics[height=1cm]{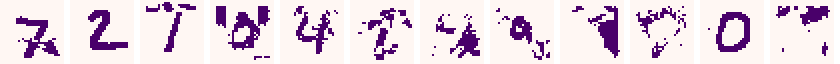} \\
\textit{SGAN-S}           & \includegraphics[height=1cm]{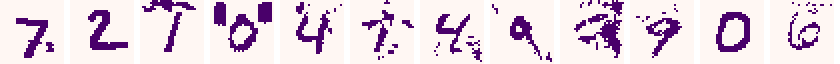} \\
\textit{SGAN-S Uncond.}    & \includegraphics[height=1cm]{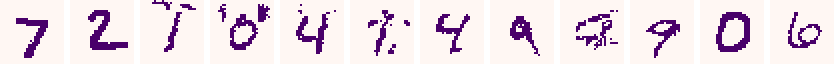} \\
\textit{SGAN-S In. Cond.}  & \includegraphics[height=1cm]{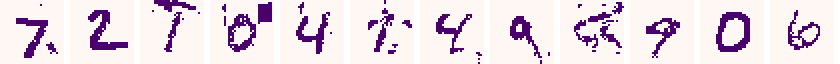} \\
\textbf{\textit{SGAN-S Out. Cond.}} & \includegraphics[height=1cm]{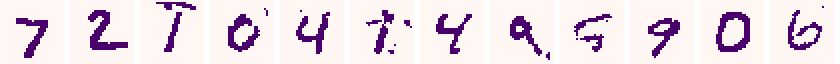}
\end{tabular}
    \caption{Segmentation samples from all models, including baselines (FCN Segmenter, SGAN-S), alternative feature discriminators (SGAN-S Uncond., SGAN-S In. Cond.), as well as our model (SGAN-S Out. Cond).}
    \label{fig:res:segm}
\end{figure*}

Table~\ref{tab:results} reports quantitative results and Figure~\ref{fig:res:segm} shows qualitative results for all methods.

\begin{table*}[ht]
\label{tab:results}
\centering
\begin{tabular}{@{}lrrrrrr@{}}
\toprule
 \multicolumn{7}{c}{\textbf{MNIST-Thin to MNIST-M}} \\
                                             & \multicolumn{1}{l}{\textbf{Src. IoU back}} & \multicolumn{1}{l}{\textbf{Src. IoU digit}} & \multicolumn{1}{l}{\textbf{Src. mIOU}} & \multicolumn{1}{l}{\textbf{Tgt. IoU back}} & \multicolumn{1}{l}{\textbf{Tgt. IoU digit}} & \multicolumn{1}{l}{\textbf{Tgt. mIOU}} \\ \midrule
\textit{FCN Segmenter}                           & 1                                            & 1                                             & 1                                        & 0.868                                        & 0.396                                         & 0.632                                    \\
\textit{SGAN-S}                   & 1                                            & 1                                             & 1                                        & 0.901                                        & 0.523                                         & 0.712                                    \\
\textit{SGAN-S Uncond.}      & 1                                            & 1                                             & 1                                        & 0.936                                        & 0.622                                         & 0.779                                    \\
\textit{SGAN-S In. cond.}  & 1                                            & 1                                             & 1                                        & 0.922                                        & 0.602                                         & 0.762                                    \\ 
\textit{\textbf{SGAN-S Out. cond.}} & 1                                            & 1                                             & 1                                        & \textbf{0.953}                               & \textbf{0.723}                                & \textbf{0.838}                           \\ \bottomrule
\end{tabular}
\vspace{0.2cm}
\caption{Quantitative results: test results on source (MNIST-thin) and target (MNIST-M) domain. Results are reported in terms of per class IoU (background and digit) as well as mean IoU (mIoU).}
\end{table*}

We first stress the results from the \textit{FCN segmenter} and \textit{StarGAN with segmenter (SGAN-S)} baselines described in Section~\ref{sec:exp:base}. None of those baselines include any domain adaptation step, and their performance on the target domain serves solely as lower bound to the unsupervised domain adaptation models. Analogously, the performance of the \textit{FCN segmenter} on the source domain may serve as upper bound for the source domain to the rest of the models. We also trained the same segmenter on the target domain to provide an upper bound on it; achieving a $0.998$ digit IoU, a background IoU of $1$ and a mean IoU of $0.999$. As expected, both baselines \textit{FCN segmenter} and \textit{SGAN-S} exhibit the lowest results, especially when it comes to segmenting the target domain digits, outlining the need of including domain adaptation constraints to the model. However, adding an image translation backbone significantly boosts the results adding $0.127$ points of digit IoU and $0.08$ points of mean IoU. The differences among \textit{FCN segmenter} and \textit{SGAN-S} can also be visually perceived in the qualitative results shown in Figure~\ref{fig:res:segm}, where some digits that were missed by the \textit{FCN segmenter}, appear in the \textit{SGAN-S} segmentations.

Second, we analyze the results of the often adopted domain adaptation strategy denoted as \textit{SGAN-S Uncond.} (for unconditioned discriminator) in Table~\ref{tab:results} and Figure~\ref{fig:res:segm}. The is \textit{SGAN-S Uncond.} model is a natural alternative to our model, which uses a global discriminator to match features from both domains at the image translation bottleneck (see Section~\ref{sec:exp:abl:condis}). We can also see \textit{SGAN-S Uncond.} as an extension of \textit{SGAN-S}, which includes a global feature matching as domain adaptation step. Following the reported results, we can see that the unconditional feature matching provides a notable improvement over the baselines. The target digit IoU increases almost $0.10$ points and the target mIoU $0.067$ points w.r.t. \textit{SGAN-S}, which results in a $0.226$ points target digit boost and $0.147$ mIoU points boost over the \textit{FCN segmenter} baseline. In the segmentation samples from Figure~\ref{fig:res:segm}, this improvement is mainly perceived by the diminishing of false positives, leading to much cleaner segmentations (but with still room for improvement).

Third, we analyze the results of an alternative local conditional discriminator introduced in Section~\ref{sec:exp:abl:condis}. We remind that in this version, the feature discriminator $D_f$ takes the segmentation predictions as input, by concatenating them to the image translation bottleneck features. This allows the discriminator to have information about the distributions of the classes in each domain. This model is denoted as \textit{SGAN-S In. Cond.}. Unfortunately, this method shows a drop in performance of $0.02$ points in the digit IoU over the unconditional version. We hypothetize that this performance drop is due to the differences that segmentation masks exhibit between source and target domains, i.e. source segmentations are clean and show sharp and thin digits whereas target segmentations are more noisy and are expected to show thick digits. This clear differentiation makes the task of distinguishing source and target domains trivial for the discriminator and breaks the game of the adversarial matching, showing no improvement w.r.t. a regular global adversarial feature matching. 

Finally, we compare the baselines and alternative model results against our model. We denote our model as \textit{SGAN-S + Out. Cond.}, which stands for output conditional discriminator. Recall that, in this case, the segmentation conditioning happens at the output of a pixel-wise discriminator and intends to match features per class (instead of globally). This method leads to the best results, with an improvement of $0.101$ points of target digit IoU w.r.t. \textit{SGAN-S + Uncond.}, $0.20$ points of target digit IoU w.r.t. \textit{SGAN-S} and $0.327$ w.r.t. \textit{FCN segmenter}. A similar trend can be observed for the target mIOU. Qualitatively, at the same time, the \textit{SGAN-S + Out. Cond.} segmentations show a much better appearance, exhibiting the cleanest and sharpest segmentation predictions among all results. Moreover, the segmented digits in the target domain seem to preserve a much more realistic thickness when performing domain adaptation. 

All the previously discussed evidences outline the positive impact of the contributed output conditional discriminator and support its effectiveness to properly match features representing different content, leading to improved segmentation predictions.

\section{Conclusions}
\label{sec:concl}
In this thesis, we have tackled the unsupervised domain adaptation problem for semantic segmentation. Specifically, we have focused on the shortcomings of the global feature matching performed in most domain adaptation methods and the high numbers of parameters that state-of-the-art approaches generally need. In order to solve this problem, we have designed a new method using the information from the segmentation predictions, to perform the feature matching locally and according to the per class distributions in each domain. At the same time, we have benefited from the \textit{StarGAN} architecture, which concatenates the domain label to the input, obtaining domain specific transformations without the need of having two completely separated domain-specific networks. With these procedures, we have been able to improve upon state-of-the-art approaches on a proof-of-concept task, while having notably less parameters to train and outlining the potential of the proposed approach. However, it is worth noting, our model still contains parameters that are only used during training, namely the image translation decoder network.  

A natural next step would be to test our method in larger scale datasets, such as SYNTHIA \cite{synthia} or GTA5 \cite{gta5} vs. Cityscapes \cite{cityscapes}, with the intention of analyzing the performance on more realistic environments. Additionally, we could also analyze the influence of different losses such as \textit{identity loss} as an alternative or supplementary loss, with the goal of simplifying the model or studying loss complementarities.

In further works, it would be interesting to consider adapting our method to video processing, instead of our current image-based (frame by frame) prediction model. Exploiting the sequential nature of video, we would allow the model to output time-consistent predictions, moving closer to applications in, for example, autonomous driving systems. Another potential future research direction would be extending our methodology to one shot or zero shot learning cases where, for example, the adaptation step would need to consider the presence of unseen classes between source and target domains.

\section*{Acknowledgments}
First of all, I would like to thank Yaroslav Ganin for joining the project and contributing with his great ideas and experience, and Faruk Ahmed for sharing his knowledge about the adversarial networks and reviewing all our implementation. I would also like to thank Joelle Pineau for giving me the opportunity of joining the McGill RLLab and funding me during this period. Joost van de Weijer for his feedback and review of the work, and Marc Masana for his help and advice. My family, and specially my mother, for their support during this year and my roommates and classmates from Barcelona for this both awesome and hard year. And finally, I would like to thank my supervisor Adriana, first of all for offering me the opportunity of joining the AI community of Montreal and letting me work with her in this project. But, I would also like to thank her for her hard work helping me with a really active and exceptional supervision, for taking care of my insertion in the AI research community, for teaching me so much about the appropriate research methodology (in which I had not any experience at all), for worrying about whether I was feeling comfortable in my stay and for going beyond and giving me advise for my uncertain future. Thank you.

\bibliographystyle{abbrv}
\bibliography{main}

\newpage
\clearpage
\appendices
\section{Model implementation details}
\label{apx:implementation}
In this appendix, we detail the implementation details of all networks used for the experiments of this thesis. We also provide information on the training procedure and all the necessary hyper-parameters.

\minisection{Fully convolutional network for semantic segmentation.} This network is explained in Section~\ref{sec:back-fcnss} and has been depicted in Figure~\ref{fig:segm}. The network is built with several convolutional blocks composed of a convolution (or transposed convolution when upsampling), a dropout layer \cite{dropout}, an instance normalization layer (without computation of running statistics) \cite{instancenorm} and a rectified linear unit (ReLU) activation. More specifically, it is composed of a first convolutional block of 32 channels and $3\times3$ kernel size, followed by two downsampling blocks with stride 2 and $4\times4$ kernel size that duplicate the input channels, two upsampling blocks to recover the input resolution (dividing by 2 the number of channels) and a final $3\times3$ convolutional layer followed by a sigmoid non-linearity. We apply a $0.2$ dropout in all convolutional blocks along the network during training.

The network is trained using an Adam optimizer \cite{adam} with an initial learning rate of $0.001$ and an exponential decay of $0.995$ after each epoch. The optimizer hyper-parameters are set as follows: $\beta_1=0.9$ and $\beta_2=0.999$. Training is performed on mini-batches of size $32$ for a maximum of $500$ epochs with early stopping patience of $50$ epochs (validation loss not improving).

\minisection{CycleGAN and StarGAN.} For the \textit{CycleGAN} \cite{cyclegan} and \textit{StarGAN} \cite{stargan} architectures presented in Sections~\ref{sec:back-cycgan} and \ref{sec:back-stargan} respectively, we mostly use the same implementation hyper-parameters published by the authors with slight modifications. Both models share a very similar architecture, in their generators and discriminators. The modifications that we perform result in the deletion of the bottleneck residual blocks and the reduction of the initial convolution's number of channels to $32$ in both models' generators. These modifications aim to reduce the models' capacity and adapt them to the datasets of interest to this thesis. In the case of \textit{StarGAN}, we also reduce the kernel size of the generator's convolutions to $3\times3$. The models have been depicted in Figure~\ref{fig:cyclegan:adv} (\textit{CycleGAN}) and Figure~\ref{fig:stargan:adv} (\textit{StarGAN}).

Both models are trained following the optimization hyper-parameters suggested by the authors. The \textit{CycleGAN} model is trained for $200$ epochs and the best results have been picked (following a visual criteria) for the examples shown in Figure~\ref{fig:translations}. In the case of \textit{StarGAN}, the model is trained for $200.000$ iterations (approximately $107$ epochs). 

\minisection{Our model.} The architecture in our model, as presented in Section~\ref{sec:method} and shown in Figure~\ref{fig:ourmodel:all}, is composed by several modules. The modules corresponding to the \textit{StarGAN} backbone follow exactly the same implementation detailed previously in this appendix. The segmentation decoder $S$ attached to $G_e$ consists of an upsampling path and a final convolutional layer following the \textit{FCN for segmentation} model, detailed in this appendix. The remaining module in the model is the feature discriminator $D_f$ that we introduce in this thesis. It is depicted in detail in Figure~\ref{fig:ourmodel:df} and consits of two upsampling convolutional blocks and a final $3\times3$ convolutional layer with a single output channel. The upsampling convolutional blocks include a $4\times4$ transposed convolution that upsamples the spatial resolution by 2 and outputs half of its input channels, followed by a $0.01$ slope leaky ReLU. All modules have a $0.2$ dropout layer in their convolutional blocks.

All the modules are trained with an Adam optimizer \cite{adam} with an initial learning rate of $0.0001$ and an exponential decay of $0.995$ every 1500 training iterations. The optimizer hyper-parameters are set as follows: $\beta_1=0.5$ and $\beta_2=0.999$. The discriminators $D$ and $D_f$ follow a $5/1$ training step ratio over the generator $G$ and the segmenter $S$. The loss component importance parameters are selected as follows: $\lambda_{dom}=1$, $\lambda_{cyc}=10$, $\lambda_{gp}=2$, $\lambda_{segm}=10$ and $\lambda_{dom}^f=1$; and finally, the whole model is trained for a maximum of $500$ epochs with an early stopping patience of $50$ epochs.

\minisection{StarGAN with segmenter baseline.} This baseline model from Section~\ref{sec:exp:base} follows the same architecture of our model without having a feature discriminator $D_f$. The training parameters are also the same as our model with the exception of the hyper-parameter $\lambda_gp=10$.

\minisection{Unconditional discriminator.} This model presented in Section~\ref{sec:exp:abl:condis} also follows our models' architecture, with a different discriminator $D_f$. This discriminator is depicted in Figure~\ref{fig:df:input} and is composed of two downsampling blocks, an average pooling and a fully connected layer. The downsampling blocks consist of a $4\times4$ convolution with stride $2$, duplicating the number of channels at its input. The convolution is followed by a leaky ReLU of slope $0.01$. The module is trained with a dropout probability of $0.2$.

The training setting is analogous to the one from our model, except for a $\lambda_{gp}$ of $5$.

\minisection{Input conditioned discriminator.} Analogously to the unconditional discriminator, this model also changes the discriminator $D_f$ architecture, exclusively. The feature discriminator, in this cas, is explained in Section~\ref{sec:exp:abl:condis} and depicted in Figure~\ref{fig:df:input}. The architecture of this discriminator is as follows: convolutional blocks of $4\times4$ convolutions and leaky ReLU of $0.01$ slope. In this discriminator, input features are first upsampled to the original spatial resolution by $2$ upsampling convolutional blocks that double the size and halve the number of channels at each step. Then, the segmentation mask is concatenated to the upsampled features and $4$ downsampling blocks are applied to reduce by a factor of 2 each axis of the spatial resolution, while duplicating the number of channels at each step (without taking into account the extra channels from the concatenation of the segmentation prediction). These features are finally processed by an average pooling, which is followed by a fully connected layer with a single output. Following previous pipelines, this discriminator is trained with a dropout probability of $0.2$.

The training setting is analogous to the one followed by our model, except for a $\lambda_{gp}$ of $1$.

\section{Additional examples of CycleGAN and StarGAN translations}
\begin{figure*}[!h]
    \centering
    \begin{tabular}{rl}
    Image               & \includegraphics[height=1cm]{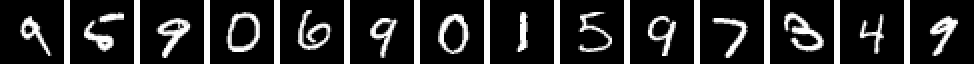} \\
    \textit{CycleGAN}   & \includegraphics[height=1cm]{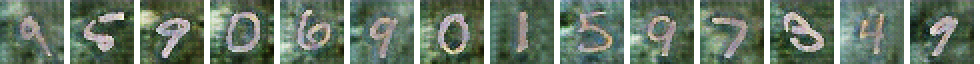} \\
    \textit{StarGAN}    & \includegraphics[height=1cm]{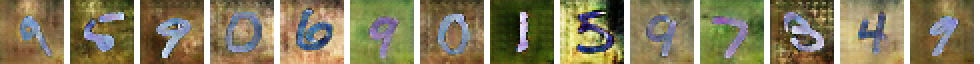}
    \\ & \\
    Image               & \includegraphics[height=1cm]{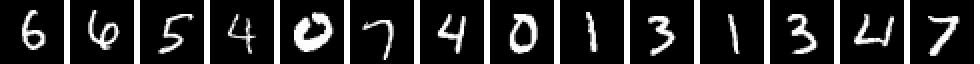} \\
    \textit{CycleGAN}   & \includegraphics[height=1cm]{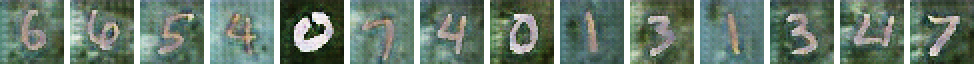} \\
    \textit{StarGAN}    & \includegraphics[height=1cm]{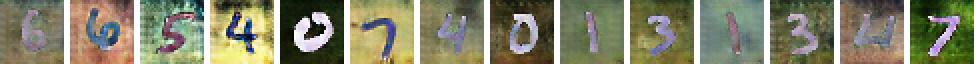}
    \\ & \\
    Image               & \includegraphics[height=1cm]{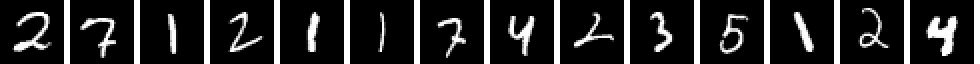} \\
    \textit{CycleGAN}   & \includegraphics[height=1cm]{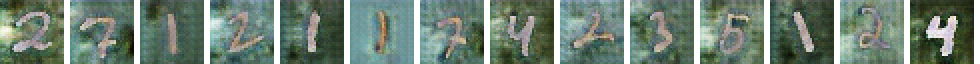} \\
    \textit{StarGAN}    & \includegraphics[height=1cm]{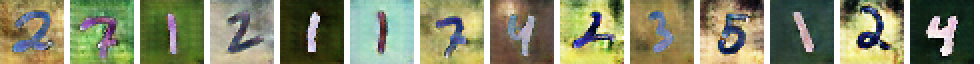}
    \\ & \\
    Image               & \includegraphics[height=1cm]{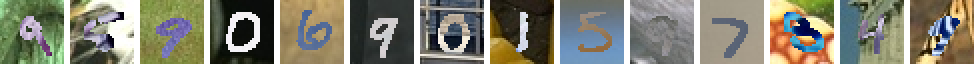} \\
    \textit{CycleGAN}   & \includegraphics[height=1cm]{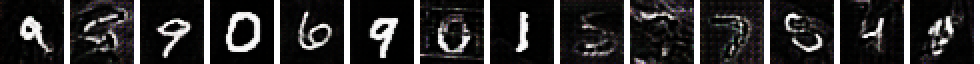} \\
    \textit{StarGAN}    & \includegraphics[height=1cm]{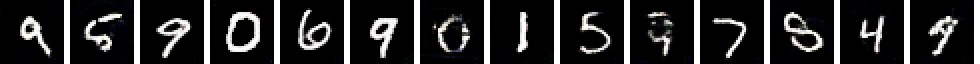}
    \\ & \\
    Image               & \includegraphics[height=1cm]{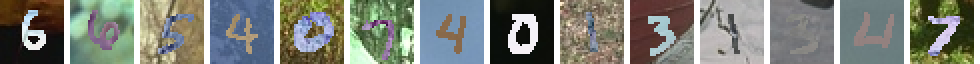} \\
    \textit{CycleGAN}   & \includegraphics[height=1cm]{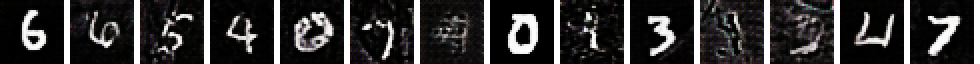} \\
    \textit{StarGAN}    & \includegraphics[height=1cm]{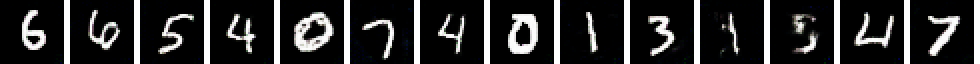}
    \\ & \\
    Image               & \includegraphics[height=1cm]{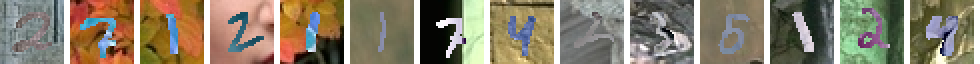} \\
    \textit{CycleGAN}   & \includegraphics[height=1cm]{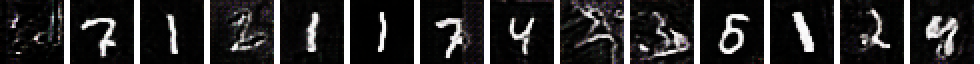} \\
    \textit{StarGAN}    & \includegraphics[height=1cm]{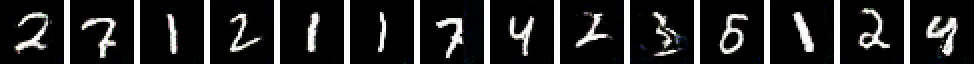}
    \end{tabular}
    \caption{Additional examples of translations between MNIST and MNIST-M.}
    \label{fig:extranslations}
\end{figure*}

In Figure~\ref{fig:extranslations} we provide additional examples of translations between MNIST and MNIST-M domains using CycleGAN and StarGAN.
\label{apx:extranslations}

\section{Aditional examples of segmentations}
\begin{figure*}[!h]
    \centering
    \begin{tabular}{rl}
    Image                               & \includegraphics[height=1cm]{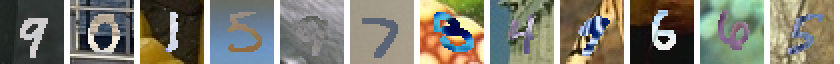} \\
    \textit{FCN Segmenter}              & \includegraphics[height=1cm]{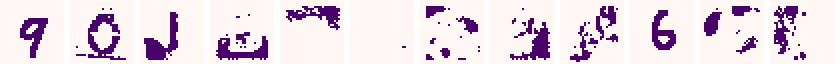} \\
    \textit{SGAN-S}                     & \includegraphics[height=1cm]{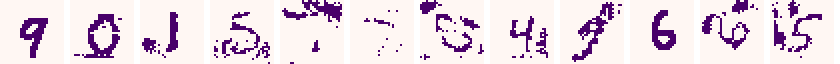} \\
    \textit{SGAN-S Uncond.}             & \includegraphics[height=1cm]{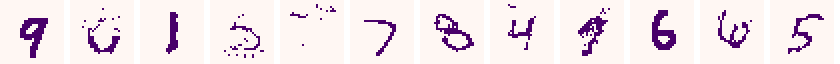} \\
    \textit{SGAN-S In. Cond.}           & \includegraphics[height=1cm]{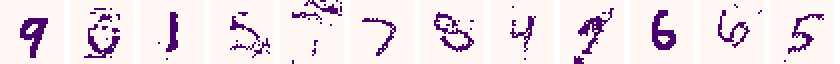} \\
    \textbf{\textit{SGAN-S Out. Cond.}} & \includegraphics[height=1cm]{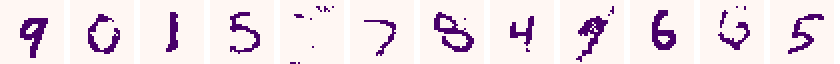} 
    \\ & \\
    Image                               & \includegraphics[height=1cm]{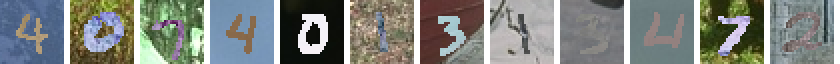} \\
    \textit{FCN Segmenter}              & \includegraphics[height=1cm]{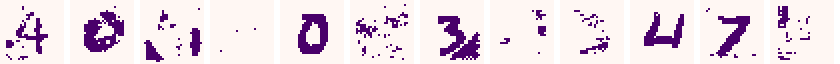} \\
    \textit{SGAN-S}                     & \includegraphics[height=1cm]{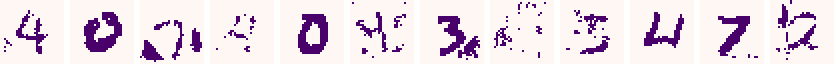} \\
    \textit{SGAN-S Uncond.}             & \includegraphics[height=1cm]{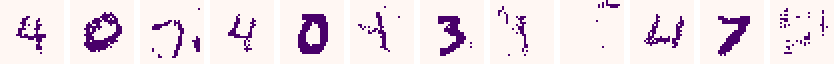} \\
    \textit{SGAN-S In. Cond.}           & \includegraphics[height=1cm]{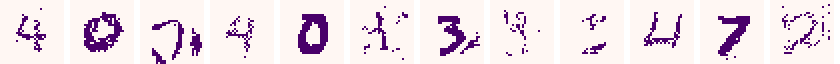} \\
    \textbf{\textit{SGAN-S Out. Cond.}} & \includegraphics[height=1cm]{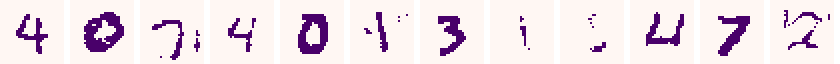} 
    \\ & \\
    Image                               & \includegraphics[height=1cm]{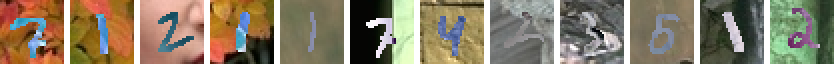} \\
    \textit{FCN Segmenter}              & \includegraphics[height=1cm]{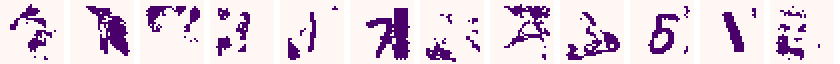} \\
    \textit{SGAN-S}                     & \includegraphics[height=1cm]{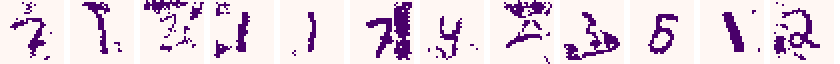} \\
    \textit{SGAN-S Uncond.}             & \includegraphics[height=1cm]{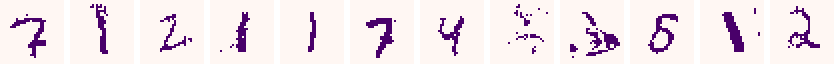} \\
    \textit{SGAN-S In. Cond.}           & \includegraphics[height=1cm]{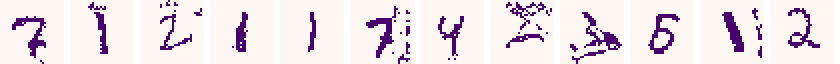} \\
    \textbf{\textit{SGAN-S Out. Cond.}} & \includegraphics[height=1cm]{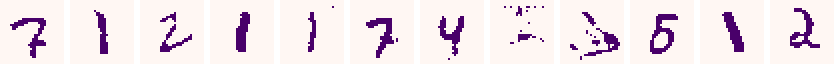} 
    \end{tabular}
    \caption{Additional segmentation samples from all models, including baselines (FCN Segmenter, SGAN-S), alternative feature discriminators (SGAN-S Uncond., SGAN-S In. Cond.), as well as our model (SGAN-S Out. Cond).}
    \label{fig:exsegmentations}
\end{figure*}

In Figure~\ref{fig:exsegmentations} we provide additional examples of segmentations using all models, including baselines (FCN Segmenter, SGAN-S), alternative feature discriminators (SGAN-S Uncond., SGAN-S In. Cond.), as well as our model (SGAN-S Out. Cond).

\ifCLASSOPTIONcaptionsoff
  \newpage
\fi

\end{document}